\documentclass[10pt,twocolumn,letterpaper]{article}

\usepackage{iccv}              %
\usepackage{graphicx}
\usepackage{multirow}
\usepackage{amssymb}

\usepackage{fancyvrb}
\usepackage{fvextra}
\usepackage{xspace}

\definecolor{iccvblue}{rgb}{0.21,0.49,0.74}
\usepackage[pagebackref,breaklinks,colorlinks,allcolors=iccvblue]{hyperref}

\usepackage{csquotes}
\usepackage{multirow}

\def\paperID{2931} %
\def\confName{ICCV}
\def\confYear{2025}

\title{SUB: Benchmarking CBM Generalization via Synthetic Attribute Substitutions}

\author{
    Jessica Bader$^{1}$\qquad
    Leander Girrbach$^{1}$\qquad
    Stephan Alaniz$^{2}$\thanks{Work was done at TUM and Helmholtz Munich.}\qquad 
    Zeynep Akata$^{1}$\\
    \small{
    $^1$Technical University of Munich, Helmholtz Munich, Munich Center for Machine Learning (MCML)
    }\\
    \small{
    $^2$ LTCI, Télécom Paris, Institut Polytechnique de Paris, France
    }
    \\
    {\tt\small jessica.bader@tum.de}
}

\definecolor{mutedgreen}{RGB}{88, 138, 102}
\definecolor{mutedred}{RGB}{163, 94, 94} 

\newcommand{\datasetname}{SUB\xspace}
\newcommand{\methodfull}{Tied Diffusion Guidance\xspace}
\newcommand{\method}{TDG\xspace}

\begin{document}
\maketitle

\begin{abstract}
Concept Bottleneck Models (CBMs) and other concept-based interpretable models show great promise for making AI applications more transparent, which is essential in fields like medicine. Despite their success, we demonstrate that CBMs struggle to reliably identify the correct concepts under distribution shifts.
To assess the robustness of CBMs to concept variations, we introduce \datasetname: a fine-grained image and concept benchmark containing 38,400 synthetic images based on the CUB dataset. To create \datasetname, we select a CUB subset of 33 bird classes and 45 concepts to generate images which substitute a specific concept, such as wing color or belly pattern. We introduce a novel \methodfull (\method) method to precisely control generated images, where noise sharing for two parallel denoising processes ensures that both the correct bird class and the correct attribute are generated. This novel benchmark enables rigorous evaluation of CBMs and similar interpretable models, contributing to the development of more robust methods.
Our code is available at 
\href{https://github.com/ExplainableML/sub}{https://github.com/ExplainableML/sub} and the dataset at \href{http://huggingface.co/datasets/Jessica-bader/SUB}{http://huggingface.co/datasets/Jessica-bader/SUB}.

.
\end{abstract}
\vspace{-20pt}
\section{Introduction}
\label{sec:intro}

While deep learning models excel on complex tasks, they are often criticized for lack of transparency in their reasoning, %
causing a severe bottleneck in the deployment of deep models in real-world contexts. For example, in the medical field, the model's reasoning must be present in order to be %
used by physicians. 
Interpretable models are essential to address these needs. One core method is the Concept Bottleneck Model (CBM)~\cite{cbm}, which generates intermediate, interpretable concepts to inform the final prediction.

\begin{figure}[t]
    \centering
    \includegraphics[width=\linewidth]{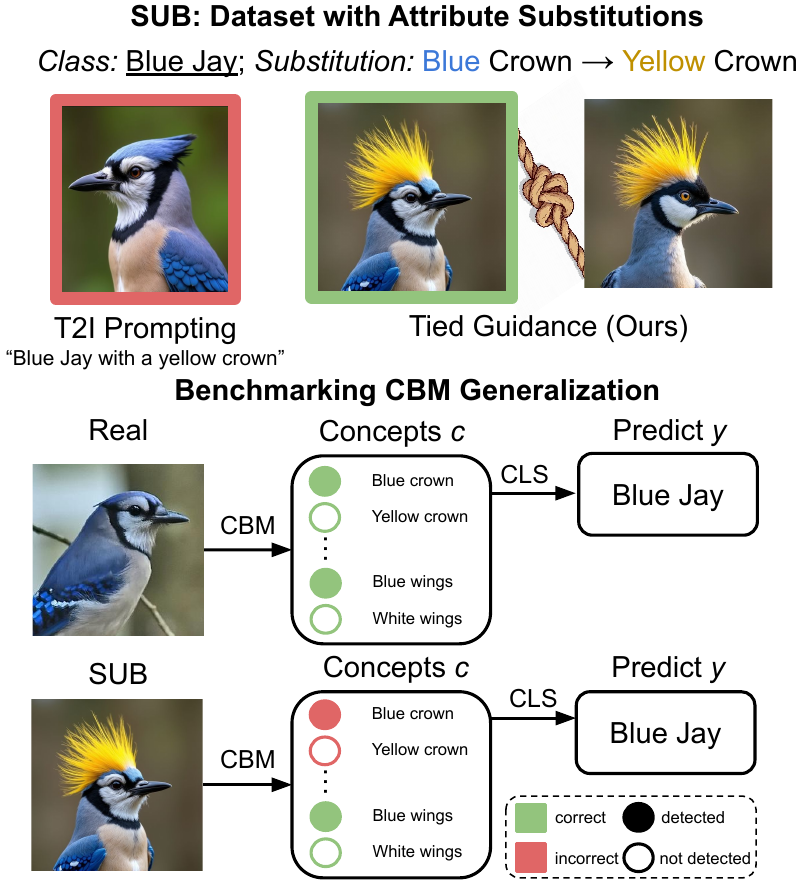} 
    \vspace{-10pt}
    \caption{
        (Top) TGD modifies attributes where %
        prompting fails. (Bottom) The CBM generalizes poorly, memorizing the ``Blue Jay'' concept vector and mis-classifying the modified concept.
    }
    \label{fig:motivation}
\end{figure}

CBM evaluation exhibits a limitation, as demonstrated in \cref{fig:motivation}. %
We expect the CBM to assign the bottom bird the label \textit{yellow crown}, but it still identifies a \textit{blue crown}. In fact, the CBM predicts the exact concept vector associated with the Blue Jay class (middle), despite clear differences. %
This observation suggests that the CBM may ground predictions in extraneous factors, potentially entirely unrelated to the visible concepts, rather than those in the image. %
Ergo, the CBM defaults to predicting the concept vector of the %
most similar training class. Such behavior raises doubts about the validity of the so-called \textit{concept predictions} as reliable interpretability tools. As \citet{cbm} evaluated their CBM solely on training classes, 
we cannot distinguish if the model genuinely learned to identify concepts or simply memorized the concept vectors of the target classes. 

In this work, our objective is to evaluate the generalization of concept predictions in CBMs and Vision Language Models (VLM) that underpin interpretable architectures, particularly when input images contain novel combinations of known concepts. More specifically, we focus on small deviations from the training classes, %
with single concept alterations. In this way, we isolate individual attributes, ensuring the model cannot rely on other cues to shortcut predictions.
Furthermore, we propose \methodfull (\method) to generate these single concept substitutions. As shown in \cref{fig:motivation} (top), naively prompting a latent diffusion model (LDM) for a \enquote{Blue Jay with a yellow crown} does not yield the desired result. Instead, \method generates a second image where \enquote{yellow crown} appears naturally. By tying the diffusion processes, we are able to successfully substitute the desired attribute on the \enquote{Blue Jay} class. %

Using \method, we create our synthetic dataset: %
\underline{S}ubstitutions on Caltech-\underline{U}CSD \underline{B}irds-200-2011 (SUB), consisting of 38,400 images for evaluating interpretable models trained on the CUB dataset~\cite{cub} or with open vocabularies. Leveraging \datasetname, we find that CBMs and VLMs fail to generalize to novel combinations of known concepts.
In particular, we show that a number of CBMs trained with both class- %
and image-level concept labels, as well as a variety of leading VLMs, cannot reliably detect %
concepts. %
This provides strong evidence that these models %
infer concepts from the predicted class rather than grounding them in the image.

In summary, our contributions are the following:
(1) We propose a test-time LDM modification to generate novel combinations of known concepts.
(2) We release \datasetname, an evaluation dataset for interpretable models, which is the first photorealistic image dataset %
to isolate concepts before evaluating classification.
(3) We reveal that existing CBMs and VLMs fail to generalize to new combinations of known concepts, raising concerns about %
their interpretability.

\section{Related Work}
\label{sec:relatedworks}
In explainable AI, concept-based models have emerged as a powerful interpretability tool, as prototypical parts~\cite{Nauta2023PIPNetPI, Chen2018ThisLL, Rymarczyk2020ProtoPSharePP, Rymarczyk2021InterpretableIC, Pach2024LucidPPNUP, van2023pdisconet, aniraj2024pdiscoformer}, sparse auto-encoders~\cite{Makhzani2013kSparseA, Cunningham2023SparseAF, Rao2024DiscoverthenNameTC, Kim2024RevelioIA, Thasarathan2025UniversalSA, Zhang2024LargeMM}, self-explaining models~\cite{AlvarezMelis2018TowardsRI}, or Concept Bottleneck Models (CBM)~\cite{cbm, Oikarinen2023LabelFreeCB, Tan2024ExplainVA, panousis2024coarse, yang2023language}. 
CBMs in particular are valued for their ability to pre-define key concepts and enable interventions, making them useful in fields like medicine~\cite{Nnamdi2023ConceptBM, Alam2024TowardsIR, Chowdhury2024AdaCBMAA}. Since their creation, CBMs have become more flexible by eliminating the need for labeled data~\cite{Oikarinen2023LabelFreeCB, yang2023language}, facilitating open-vocabulary concept addition and deletion at test time~\cite{Tan2024ExplainVA}, allowing the integration of unsupervised concepts~\cite{Sawada2022ConceptBM}, improving intervention success~\cite{Singhi2024ImprovingIE}, and more. 

Nonetheless, follow-up CBM evaluation has revealed that many do not function as intended~\cite{Margeloiu2021DoCB, Mahinpei2021PromisesAP, Heidemann2023ConceptCA, Raman2024DoCB, Sinha2022UnderstandingAE}. %
Much of this research has focused on information leakage in the pre-defined concepts~\cite{Margeloiu2021DoCB, Mahinpei2021PromisesAP, Marconato2022GlanceNetsIL, Havasi2022AddressingLI, Zarlenga2023TowardsRM}, a phenomenon that has been linked to soft labels~\cite{Mahinpei2021PromisesAP}. 
Although our work does not focus on dataset leakage,
these related works have revealed that CBMs 
tend to focus on incorrect visual cues and are 
prone to overfitting to irrelevant information.
Other CBM analyses have explored their robustness~\cite{Sinha2022UnderstandingAE}, how they respect image locality~\cite{Raman2024DoCB}, the impact of concept correlation~\cite{Raman2024UnderstandingIR, Heidemann2023ConceptCA}, and their performance on cleaner tasks~\cite{Furby2024CanWC}. Heidemann et al.~\cite{Heidemann2023ConceptCA} demonstrated that CBMs struggle with attribute classification, specifically when given highly correlated concepts. Different from the previous work, we evaluate novel combinations of known concepts,
showing that concept predictions are not grounded in the image. This may reflect limited training-time attribute combinations, indicating insufficient compositional support for generalization.~\cite{Wiedemer2023CompositionalGF}.

Tangentially, image generation models have been gaining attention for
impressive generation capabilities~\cite{flux, Podell2023SDXLIL, Rombach2021HighResolutionIS}. %
Research has focused on enhancing prompt-following~\cite{Eyring2024ReNOEO, Feng2023RanniTT, Chen2024EnhancingPF, Wang2024OnDP, sega} and improving controllability~\cite{Zhang2023AddingCC, Zhao2023UniControlNetAC, Mo2023FreeControlTS}. Efforts to support compositionality have explored the combination of models~\cite{Du2024PositionCG}, objects~\cite{Du2023ReduceRR, Liu2022CompositionalVG, Du2020CompositionalVG}, attributes~\cite{Gaudi2025CoInDEL}, relations~\cite{Liu2021LearningTC}. Composed GLIDE~\cite{Liu2022CompositionalVG} 
favors object co-occurrence (e.g., showing multiple birds) through their noise injection method, compared to \method which excels at texture edits and attribute manipulation.
While CoInD~\cite{Gaudi2025CoInDEL} handles attributes, it requires training-time integration, unlike \method, which adapts at test time.

Alongside the improved capabilities of these models, there is increasing exploration into synthetic data,
both for training~\cite{Sariyildiz2022FakeIT, He2022IsSD, Kim2024DataDreamFG, Dunlap2023DiversifyYV, Dosovitskiy2015FlowNetLO} and
evaluation~\cite{Hesse2023FunnyBirdsAS, Johnson2016CLEVRAD, Park2019RobustCC}.
Previous works have explored the use of synthetic datasets specifically to enhance explainability~\cite{Hesse2023FunnyBirdsAS, Heidemann2023ConceptCA}. Synthetic images enable precise manipulations to isolate individual features, which is particularly valuable for evaluating and enforcing explainability. FunnyBirds~\cite{Hesse2023FunnyBirdsAS} consists of imaginary birds created from independent attributes, which are removed and replaced to evaluate model explanations. Similarly, Heidemann et al.~\cite{Heidemann2023ConceptCA} tested CBMs using a comparable dataset. Different from the previous works, the images in \datasetname are far more natural, bringing them much closer to real-world problems. Since our generated images resemble the types of birds found in the widely used CUB dataset~\cite{cub}, they can be leveraged to evaluate existing models trained on bird classes, eliminating the need for specialized training and creating a natural evaluation environment.

\section{\methodfull (\method)}
\label{sec:method}
\begin{figure*}[t]
    \centering
    \includegraphics[width=\linewidth]{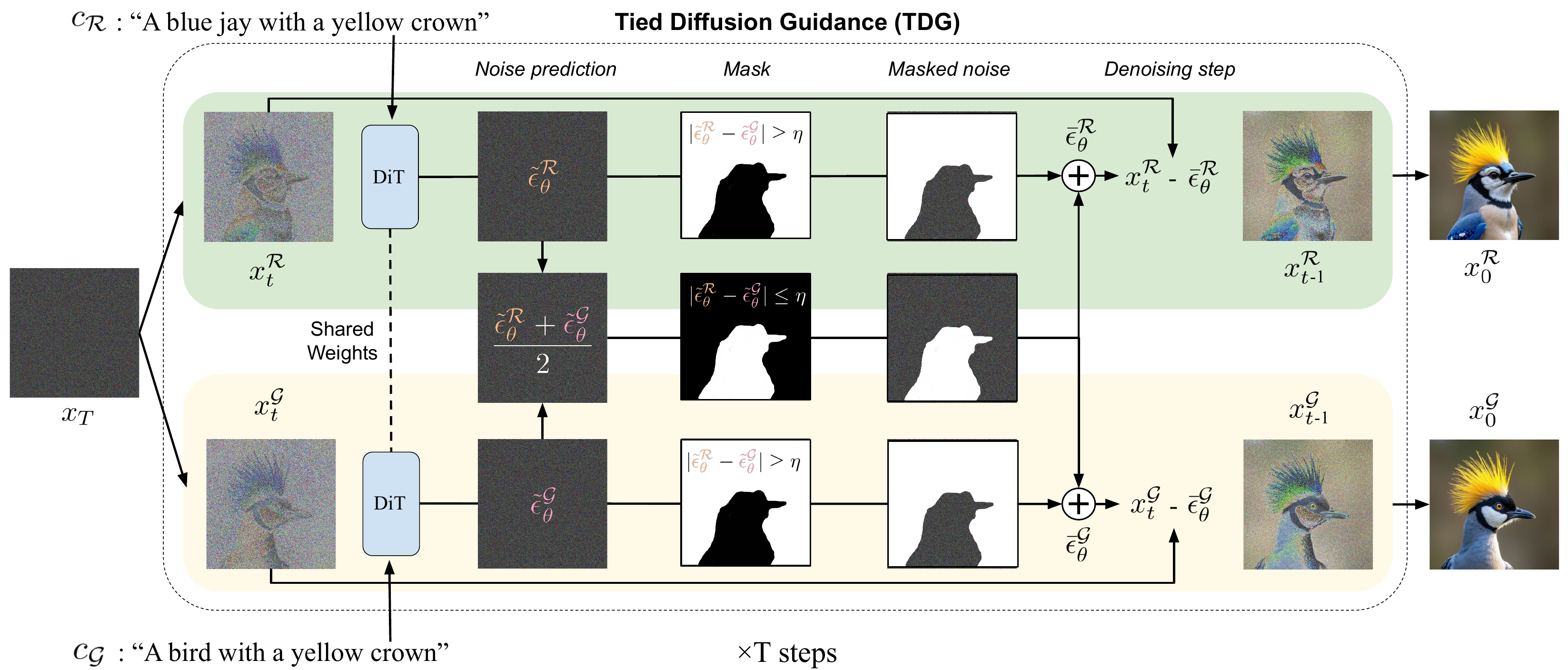} 
    \vspace{-10pt}
    \caption{%
    In Tied Diffusion Guidance, two images are generated from related prompts. At each step, the Diffusion Transformer (DiT) predicts the noises $\tilde{\epsilon}_{\theta,t}^\mathcal{R}$ and $\tilde{\epsilon}_{\theta,t}^\mathcal{G}$ for each image. We compare these predictions and, thresholded by $\eta$, retain the original noise where they differ and average the predictions where they are similar. The modified noises $\bar{\epsilon}_{\theta,t}^\mathcal{R}$ and $\bar{\epsilon}_{\theta,t}^\mathcal{G}$ are subtracted from the images, denoising them. This is repeated for $T$ steps with decreasing $\eta$, ensuring the images are highly constrained at the start but independent by the end.
    }
    \label{fig:method}
\end{figure*}

To build our SUB benchmark, we make fine-grained edits of individual bird attributes from the CUB dataset~\cite{cub}.
While LDMs follow prompts well, they falter on novel attribute combinations. To address this, we propose Tied Diffusion Guidance (TDG), a test-time method that enhances attribute-level control in text-to-image LDMs.

\textbf{Latent Diffusion Models (LDMs)} generate images by denoising a sample from Gaussian noise $x_T$ to a target image $x_0$ over $T$ steps.
For text-to-image tasks, they model $p(x|c)$, where $c$ is a text prompt. In practice, $\epsilon_{\theta}$ is trained to predict the noise in $x_t$ given $c$ at each step $t$, using the loss:
\begin{equation}
    \min_{\theta} \,\, \mathbb{E}_{(x,c) \sim \mathcal{D}, \, \epsilon \sim \mathcal{N}(0,1), \, t} \, \left[\, \left\| \, \epsilon - \epsilon_{\theta} (x_t, c, t) \, \right\|_2^2 \,\right] \, .
\end{equation}

\textbf{Diffusion Guidance.}
The LDM is trained both with text conditions and unconditionally ($c=\varnothing$). Classifier-free guidance~\cite{ho2022classifier} is applied at inference time using:
\begin{equation}
    \tilde{\epsilon}_{\theta}(x_t,c,t) = \epsilon_{\theta}(x_t,t) + s_g \left( \epsilon_{\theta}(x_t,c,t) - \epsilon_{\theta}(x_t,t) \right)
\end{equation}
where guidance scale $s_g$ controls prompt condition strength.
While text-to-image models generally follow prompts well, they often fail at zero-shot compositions, for example, ignoring the \enquote{yellow crown} edit in favor of a typical \enquote{Blue Jay} (\cref{fig:motivation}, top left). As prompts alone do not capture all \datasetname edits, we propose a test-time adaptation with semantic guidance from a reference image for attribute modifications.

\textbf{\methodfull (\method).}
Our goal is to generate an image of a \textit{reference} class $\mathcal{R}$ with an \textit{attribute substitution} $\mathcal{S}$, replacing the original attribute $\mathcal{S}^-$ (e.g., blue crown) with a target attribute $\mathcal{S}^+$ (e.g., yellow crown), while preserving $\mathcal{R}$'s remaining attributes. To overcome LDMs' struggles with zero-shot composition, we introduce guidance from a \textit{guidance} class $\mathcal{G}$, where the target attribute $\mathcal{S}^+$ is in-distribution and easier to generate. As shown in \cref{fig:method}, we propose to tie the generation of the two images, 
using $\mathcal{G}$ to guide $\mathcal{R}$ to have one attribute substituted according to $\mathcal{S}$.

To achieve single-attribute substitution, we generate paired images with separate prompts, $c_\mathcal{R}$ and $c_\mathcal{G}$, related to $\mathcal{R}$ and $\mathcal{G}$ \textit{with target attribute} $\mathcal{S}^+$, respectively. We start from the same noise $x_T^\mathcal{R} = x_T^\mathcal{G}$, and tie the noise predictions element-wise.
Given two independent noise predictions $\tilde{\epsilon}^{(1)}$ and $\tilde{\epsilon}^{(2)}$, we apply
\begin{multline}
    \mu(\tilde{\epsilon}^{(1)}, \tilde{\epsilon}^{(2)}, \eta)_i = \\
    \begin{cases}
        \frac{\tilde{\epsilon}^{(1)}_i + \tilde{\epsilon}^{(2)}_i}{2} & \text{where } |\tilde{\epsilon}^{(1)}_i - \tilde{\epsilon}^{(2)}_i| \leq \eta^{\text{th}} \text{ percentile} \\
        \tilde{\epsilon}^{(1)}_i & \text{otherwise}
    \end{cases}
\end{multline}
to obtain the mean noise prediction for all elements $i$ (i.e. image pixels) where the prediction difference is below the $\eta^{\text{th}}$ percentile and keep the noise prediction $\tilde{\epsilon}^{(1)}$ otherwise. %
We define an $\eta$ schedule that begins with the two predictions strongly tied and is loosened towards the end of generation:
\begin{equation}
    \eta(t, t_{\text{min}}, t_{\text{max}}, k) = \begin{cases}
        1 & \text{if } t > t_{\text{max}}\\
        \left(\frac{t -t_{\text{min}}}{t_{\text{max}} - t_{\text{min}}}\right)^{k} & \text{if } t_{\text{min}} \leq t \leq t_{\text{max}} \\
        0 & \text{if } t < t_{\text{min}}
    \end{cases}
\end{equation}
where $t_{\text{max}}$ controls the %
length of the initial strict noise tying phase, and $k$ regulates
the transition to
independent generation (from $t_{\text{min}}$ onwards).

\begin{figure*}
\centering
\includegraphics[width=\linewidth]{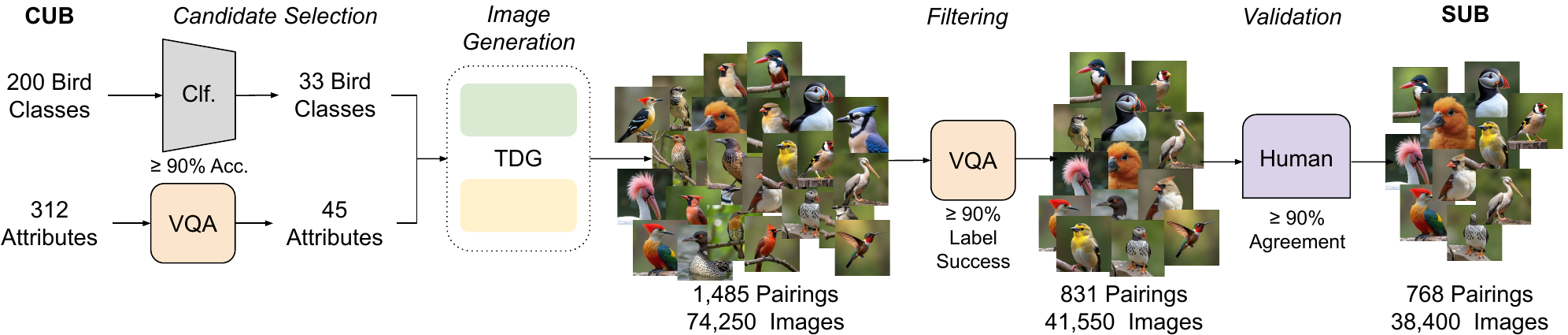}
\caption{
Meticulous filtering ensures that \datasetname is faithful to the target classes and attributes. Starting with the CUB label set~\cite{cub}, we retain only the best images from the most consistent and detectable bird-attribute pairings.
}
\label{fig:pipeline}
\end{figure*}

We apply noise tying symmetrically to both images and
update their individual noise predictions using:
\begin{equation}
    \bar{\epsilon}_{\theta,t}^\mathcal{R} = \mu(\tilde{\epsilon}_{\theta,t}^\mathcal{R}, \tilde{\epsilon}_{\theta,t}^\mathcal{G}, \eta(t, t_{\text{min}}, t_{\text{max}}, k))
\end{equation}
\begin{equation}
    \bar{\epsilon}_{\theta,t}^\mathcal{G} = \mu(\tilde{\epsilon}_{\theta,t}^\mathcal{G}, \tilde{\epsilon}_{\theta,t}^\mathcal{R}, \eta(t, t_{\text{min}}, t_{\text{max}}, k))
\end{equation}
where $\tilde{\epsilon}_{\theta,t}^\mathcal{R} := \tilde{\epsilon}_\theta(x_t^\mathcal{R},c_\mathcal{R}, t)$ and $\tilde{\epsilon}_{\theta,t}^\mathcal{G} :=\tilde{\epsilon}_\theta(x_t^\mathcal{G},c_\mathcal{G}, t)$.
In summary, \method ties noise predictions for similar pixels but allows divergence where prompt guidance differs. During generation, we gradually loosen the constraint, ultimately generating the images independently with $c_\mathcal{R}$ and $c_\mathcal{G}$, and %
ultimately
discarding the guide image.

\section{\datasetname~Dataset}
\label{sec:dataset}

We introduce \datasetname, a benchmark with fine-grained attribute edits to evaluate concept prediction faithfulness. SUB
builds upon CUB~\cite{cub}, comprising 38,400 synthesized images of bird classes with substituted attributes generated with \method. It consists of 768 unique combinations of CUB species $\mathcal{R}$ and target attributes $\mathcal{S}^+$, with 50 images per combination. The creation process is detailed in this section and visualized in \cref{fig:pipeline}. 

\textbf{CUB Preliminaries.}
The CUB dataset~\cite{cub} consists of 11,788 images of 200 bird species, annotated with attributes about bird parts and their properties. It is commonly used for fine-grained classification and model explainability.
CUB includes 28 attribute groupings $\mathcal{A}$ (e.g., leg color, belly pattern), where each image is labeled with 312 binary values representing the presence or absence of individual attributes (e.g., black leg, spotted belly).

\subsection{Prompts}
\label{sec:prompts}
We generate synthetic images using our novel \method (described in Section \ref{sec:method}) with the state-of-the-art (SOTA) text-to-image DM FLUX.1-dev~\cite{flux}. Attribute substitutions $\mathcal{S}$ are chosen from three categories: color, shape, and pattern, where a target attribute $\mathcal{S}^+ \in \mathcal{A}_\mathcal{S} \setminus \{\mathcal{S}^-\}$ is selected to differ from the reference bird. As the three categories vary in difficulty, prompts are adjusted to achieve the desired modification and avoid attribute ambiguity, requiring a minimal amount of human intervention. The full details on reference birds $\mathcal{R}$, guidance birds $\mathcal{G}$, substitutions $\mathcal{S}$, and example prompts $c_\mathcal{R},c_\mathcal{G}$ are found in Appendices A, B, and D.

\textbf{Color} is the easiest substitution for FLUX, allowing the generic term $\mathcal{G} = $ ``\textit{bird}''. We use $c_\mathcal{R} = $ ``\textit{a photo of a} $\{\mathcal{R}\}$ \textit{with a} $\{\mathcal{S}^+\}$'' and $c_\mathcal{G} = $ \textit{a photo of a bird with a } $\{\mathcal{S}^+\}$.

\textbf{Shape} is of medium modification difficulty, hence we explicitly specify the guide bird (e.g. $\mathcal{G}_{\text{cone beak}} = $ ``\textit{song sparrow}''). We discourage excessive modifications by encouraging the retention of the reference bird's body shape, using $c_\mathcal{R} = $ ``\textit{a photo of a} $\{\mathcal{R}\}$ \textit{with the body of a} $\{\mathcal{R}\}$ \textit{and a beak like a} $\{\mathcal{G_{\mathcal{S}^+}}\}$'' and $c_G =$ ``\textit{a photo of a} $\{\mathcal{G_{\mathcal{S}^+}}\}$''.

\textbf{Texture} is the toughest substitution, %
also requiring a specific guidance bird $\mathcal{G}_{\mathcal{S}^+}$ per target attribute $\mathcal{S}^+$. $\mathcal{S}^+$ is included in $c_\mathcal{G}$ %
to ensure $\mathcal{S}^-$ from $\mathcal{R}$ is changed to match $\mathcal{S}^+$ from $\mathcal{G}$, rather than vice versa. We use $c_\mathcal{R} =$ ``\textit{a photo of a} $\{\mathcal{R}\}$ \textit{with a} $\{\mathcal{S}^+\}$ \textit{like a} $\{\mathcal{G}_{\mathcal{S}^+}\}$'' and $c_\mathcal{G} =$ ``\textit{a photo of a} $\{\mathcal{G}_{\mathcal{S}^+}\}$ \textit{with a} $\{\mathcal{S}^+\}$''.

\begin{figure*}[t]
    \centering
    \includegraphics[width=\linewidth]{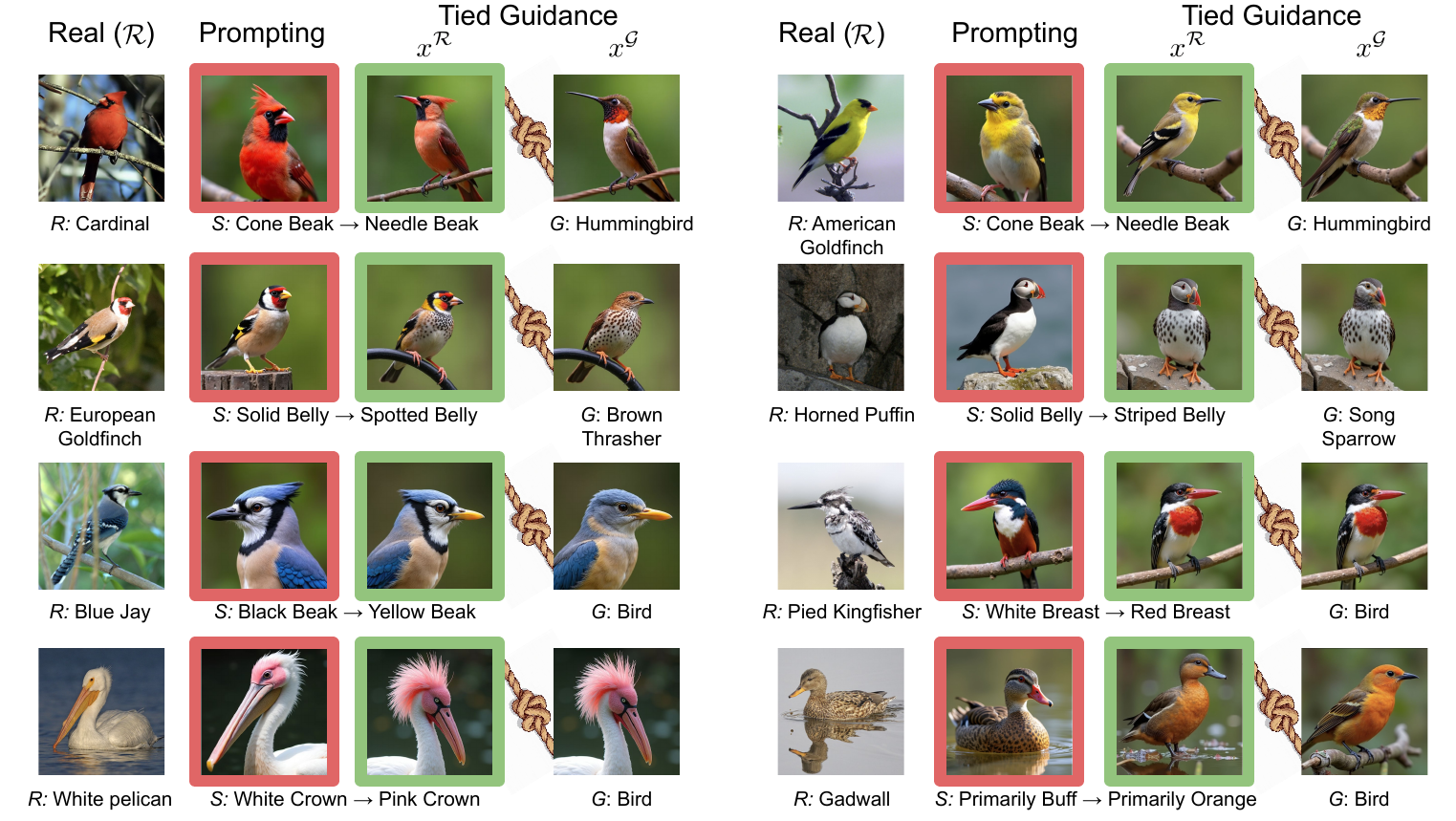} 
    \vspace{-10pt}
    \caption{
         Images \color{mutedgreen} generated with \method (green) \color{black} are high-quality and more faithfully represent both the reference bird $\mathcal{R}$ and target attribute substitution $\mathcal{S}$ than those \color{mutedred} generated with prompting alone (red) \color{black}. \method generates an additional guidance image $x^\mathcal{G}$, which we discard.
    }
    \label{fig:qualitative}
\end{figure*}

\subsection{Data Filtering}
\label{sec:filtering}

We use a filtering mechanism, common in synthetic dataset creation~\cite{Dunlap2023DiversifyYV, He2022IsSD}, to ensure the faithful representation of manipulated attributes. %
First, we identify suitable candidates for reference birds and attribute substitutions and generate $N$ images for each bird-attribute pairing. Next, we use an automatic visual-question-answering (VQA) evaluation, together with a human validation step to identify and remove images that did not modify $\mathcal{S^+}$ correctly or deviate from the reference bird $\mathcal{R}$. This process helps quantify the added variety and reliability of our attribute modifications.

\textbf{Candidate Selection.} We start by identifying suitable reference birds and attributes. Reference birds must be reliably depicted by the generative model. Hence, we generate 20 images per CUB class with FLUX~\cite{flux} and measure per-class classification accuracy with a CUB pre-trained model\footnote{\url{https://huggingface.co/Emiel/cub-200-bird-classifier-swin}} which achieves 88\% accuracy on the CUB test set.
We select the 33 classes for which the classifier achieves 100\% accuracy on the listed in %
Appendix A). %

For attribute candidates, we evaluate a VQA model's ability to identify them by
creating questions for the 28 attribute groups $\mathcal{A}_i$ with the prompt
\begin{Verbatim}[breaklines=true, breakanywhere=true,commandchars=\\\{\},codes={\catcode`$=3\catcode`_=8}]
What type of $\{\mathcal{A}_i\}$ does this bird have?
Please pick between
A) $\{a_1\}$
A) $\{a_2\}$
C) $\{\dots\}$
D) Other
\end{Verbatim} 
where $\{\mathcal{A}_i\}$ corresponds to the $i$-th attribute group (e.g. \enquote{eye color}) and the options $a_j \in \mathcal{A}_i$ contain all manifestations of that attribute group (e.g. ``red'', ``black'', etc). We also include \enquote{Other}, 
for when none of the attributes match. 
As model, we use InternVL-2.5-8B \cite{chen2024expanding,chen2024far,chen2024internvl},
which yields the best performance-efficiency tradeoff.
We run the VQA evaluation on the full CUB dataset and choose the 45 attributes that obtain accuracy $\geq 90\%$ when optimizing the answer probability threshold between $60\%$ and $95\%$.\\
\indent\textbf{VQA Filtering.}
We construct substitutions $S$ by pairing attribute candidates $\mathcal{S}^+$ with bird candidates $\mathcal{R}$ that do not already have this attribute ($\mathcal{S}^+ \neq \mathcal{S}^-$), resulting in 1,485 bird-attribute combinations. We generate images for these combinations using \method and verify the correctness of $\mathcal{S}^+$ with a VQA model (InternVL-2.5-8B) using the same prompt as attribute candidate selection. Images with incorrect attribute predictions are discarded, and we rank those remaining by target answer confidence (i.e., probability). We retain only the top 10\% of images, corresponding to 50 out of 500 images generated per pairing. If filtering retains less than 10\%, the pairing is eliminated. After this stage, 831 bird-attribute pairings remain for the \datasetname dataset.

\subsection{Human Validation}
\label{sec:human_val}

To assess the quality of our synthetic images, we conduct a human study to verify the per-attribute VQA accuracy and the faithfulness of attribute modifications $\mathcal{S}^+$ to the reference bird classes $\mathcal{R}$. Human annotators evaluate 40 randomly selected images per attribute, answering whether the attribute modification was faithfully applied and whether the generated bird deviates significantly from the reference bird. For both questions, there are three possible responses: "yes", "somewhat", and "no" (the human annotation interface is shown in %
Appendix E). We discard candidate attributes with $< 90\%$ accuracy across all reference birds, based on annotators choosing "yes". Additionally, we remove individual bird-attribute pairings with low attribute accuracy, excessive modifications beyond the target attribute, or where the reference birds already displayed the target attribute ($\mathcal{S}^+ = \mathcal{S}^-$). The resulting \datasetname dataset consists of 33 CUB bird classes, 15 attribute groupings with 32 unique target attributes, and 768 unique bird-attribute pairings, totaling 38,400 images (50 images per pairing).

\section{Experiments}
\label{sec:experiments}

In this section, we first examine the quality of our \datasetname dataset, followed by an evaluation of CBM models on \datasetname to assess their performance independently of the reference bird's appearance.
\datasetname's images are generated with FLUX.1-dev~\cite{flux} with default guidance scale 3.5. For \method, we use $k=10$, $t_{\text{min}} = 0.2$, $t_{\text{max}} = 0.6$ for pattern attributes, and $t_{\text{max}} = 0.9$ for shape and color attributes. %
These values were selected by looking at 3-5 birds across 2-4 specific attributes for each category (color, texture, shape). We generated 50 images per pair for all $33 \times 45$ bird-attribute pairs, and checked that 5+ images per pair for 10+ birds were suitable on 2-3 attributes. Optimizing the hyperparameters across the full pipeline would be too costly.

\subsection{Qualitative \method Examples from \datasetname} Figure~\ref{fig:qualitative} illustrates eight examples generated with \method. For each image, we include the real reference bird $\mathcal{R}$, and a sample generated by prompting FLUX~\cite{flux} out-of-the-box to produce the target substitution $\mathcal{S}^+$, using the same prompt as the reference image $x^\mathcal{R}$ from \method (exact prompts are described in Section~\ref{sec:prompts}).
Finally, we present both the reference image $x^\mathcal{R}$ from \datasetname, and the corresponding guidance image $x^\mathcal{G}$.
From these examples, we observe that \method is capable of generating realistic attribute-modified images. \method reliably alters the target attribute's shape, color, or texture while maintaining the overall recognizability of the reference bird, as seen in the top-left image of the Cardinal with a needle-shaped beak. While standard diffusion generates simple modifications successfully, like changing the White Pelican's crown to pink (bottom left), \method is capable of modifying more challenging attributes such as texture (second row). Interestingly, although the resulting guidance bird image $x^\mathcal{G}$ is usually visually distinct from $x^\mathcal{R}$, it sometimes converges to the same image (bottom left, third row on the right). Despite this, \method remains essential for applying target substitutions.

\begin{figure}[t]
    \centering
    \includegraphics[width=\linewidth]{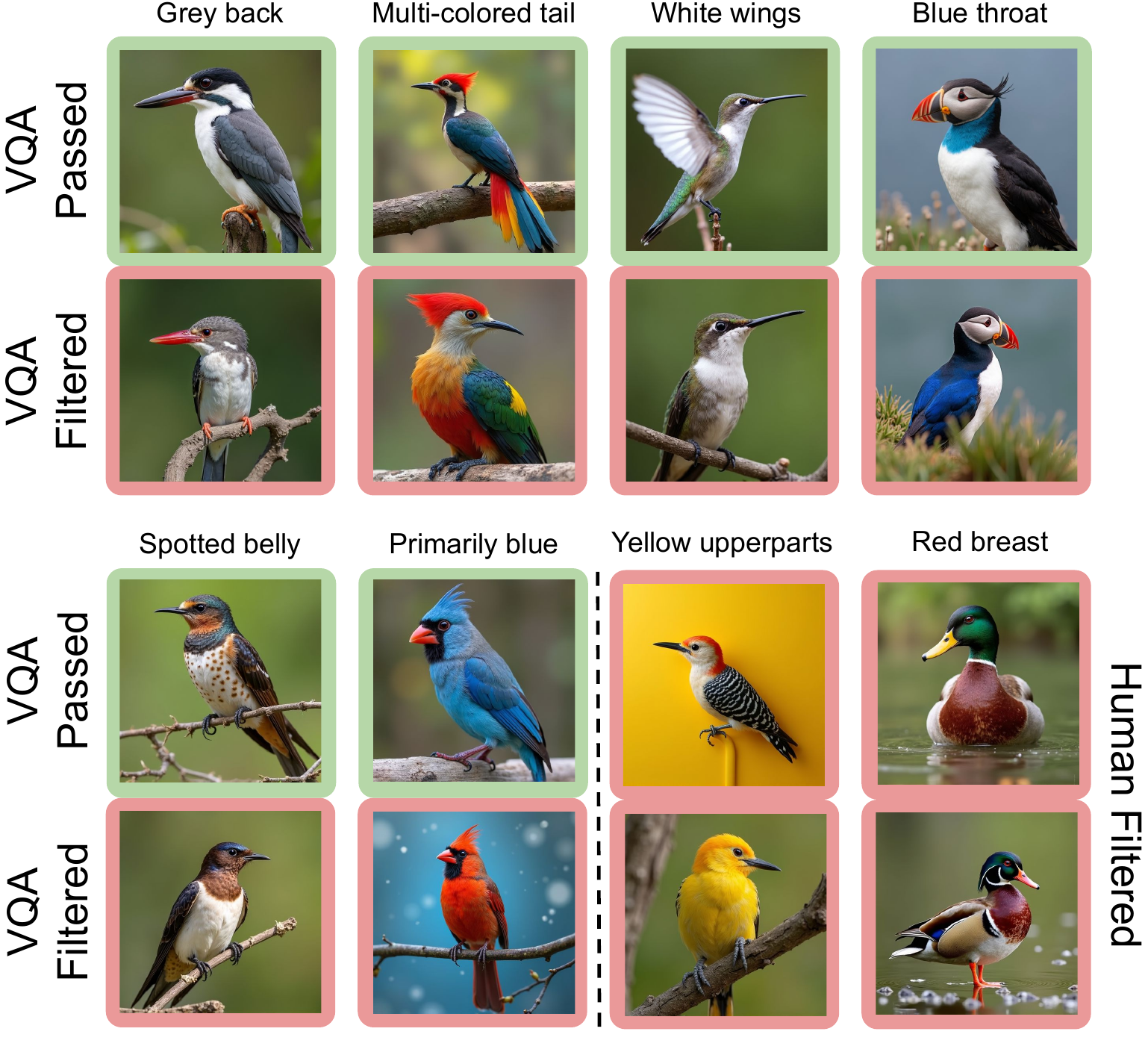} 
    \vspace{-10pt}
    \caption{
        Many \color{mutedred} failure cases \color{black} are flagged by our VQA filtering \color{mutedred} (red)\color{black}, leaving \color{mutedgreen} well-modified images \color{black} in \datasetname \color{mutedgreen}(green)\color{black}.
    }
    \label{fig:filtering}
\end{figure}

\begin{figure*}
\centering
\includegraphics[width=\linewidth]{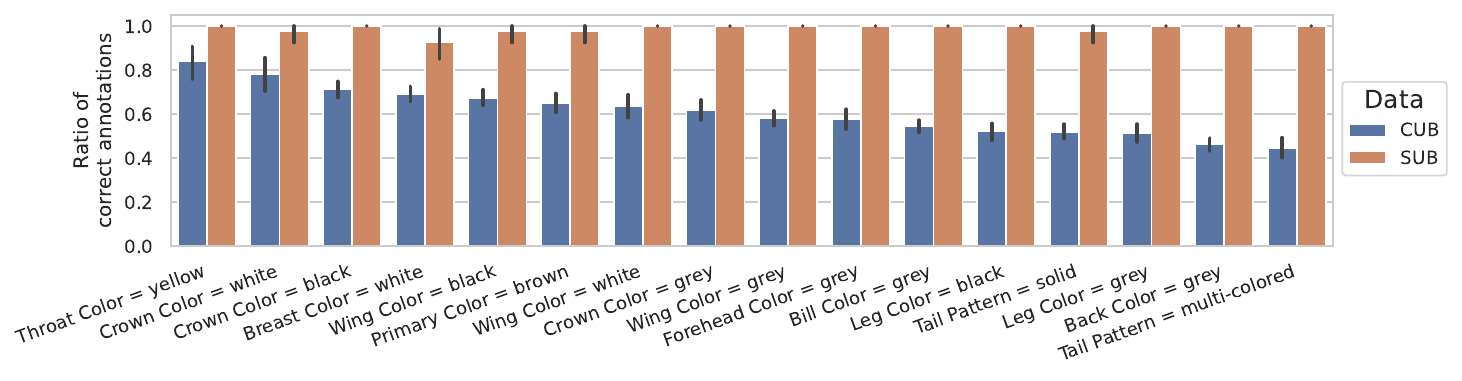}
\caption{
For all 17 attributes included in both \datasetname and CUB~\cite{cub}, \datasetname has higher annotation accuracy (std. as error bar). For CUB, we measure agreement between image-level and class-level attributes. For \datasetname, we use the attribute score from our human validation study.
}
\label{fig:entropy}
\end{figure*}

\subsection{VQA and Human Filtering Results}
VQA filtering eliminates images where \method substituted improperly. In \cref{fig:filtering}, we present results for errors flagged by our filtering method and images that passed.
InternVL-2.5-8B filters images where the target attribute is not visible (top left), the wrong attribute is modified (top right), or the model does not make meaningful modifications (bottom left). Through human verification, we further find and remove %
bird-attribute combinations where the VQA model is unreliable. The images contained in \datasetname (green) are high quality and consistently show the target attribute.

\subsection{SUB vs. CUB Label Correctness}
Although widely used, CUB~\cite{cub} has faced criticism for labeling errors~\cite{Horn2015BuildingAB}. To address inconsistent labeling, CBMs are trained with per-class attribute vectors determined by majority voting~\cite{cbm}. We ensure that \datasetname does not inherit these inconsistencies by evaluating attribute quality with human annotations.
For CUB, we group images by the presence of an attribute label in the per-class attribute vector and calculate the ratio of images where the per-image label matches (i.e. individual annotations align with the class concepts).
For \datasetname, we calculate average score (\enquote{yes} $=$ 1, \enquote{somewhat} $=$ 0.5, \enquote{no} $=$ 0) per attribute across the images in the human validation study (see \cref{sec:filtering}).
The results for 17 attributes (overlap of 32 \datasetname attributes and 112 attributes typically used for CBMs) are presented in \cref{fig:entropy}.
As previously mentioned, CUB image-level annotations are often inconsistent with class-level attribute vectors, with only 57.50\% of labels agreeing with the class-level annotations.
On the other hand, \datasetname's annotations accurately represent the images, with 98.90\% of all attributes being correctly labeled according to human agreement.

\subsection{CBM Evaluation Setup}
Using \datasetname as an evaluation set, we assess CBMs ability to generalize to our novel combinations of known concepts. 

\textbf{Base CBMs.} CBMs can be trained in three ways~\citep{cbm}:
1) jointly, where the image-to-concept and concept-to-label components are trained simultaneously using supervised learning;
2) sequentially, where the image-to-concept component is trained first, then the concept-to-label component with the former frozen; and
3) independently, where the two components are trained separately, with the class-to-label network receiving only ground-truth concept labels.
We exclude sequential training from our evaluation, as it yields the same concept predictions as independent training.

\textbf{Per-Concept CBM.} CBMs can be trained individually per concept to improve concept accuracy~\cite{Heidemann2023ConceptCA}, making the model less likely to look at spurious information, as it does not share weights with other attributes.

\textbf{Concept Embedding Models (CEM).} CEMs attempt to overcome the interpretability-performance tradeoff seen in CBMs by using a concept vector rather than a binary prediction~\cite{Zarlenga2022ConceptEM}. This vector contains two activations, one symbolizing a concept's presence and the second its absence.

\textbf{Concept Labels.} We explore labeling choices. The CBM authors proposed to train the model with fixed attribute vectors per class~\cite{cbm}. However, fixing the labels across images could hurt generalization by forcing the model to output identical predictions regardless of inter-image fluctuations. We explore models trained with class-defined attribute vectors and with the original per-image labels. Moreover, concepts can be supervised by binary attribute labels (hard) or by annotator confidence (soft). 

\textbf{Human Labeling: \datasetname.} To ensure \datasetname's high quality, we establish a human baseline where volunteers label our modified attributes. For each bird-attribute pairing, we test three images and prompt humans to select which attribute within the target group is present. We then calculate the accuracy of the human labels with respect to our intended labels. The User Interface is shown in Appendix E. %

\textbf{Human Labeling: CUB.} 
We calculate the accuracies on the original CUB classes by generating class-level labels through majority voting across all images in each bird class. We then compute the proportion of image-attribute labels where the image-level annotation agrees with the class-level majority. This helps assess how well the class-level majority approximates the image-level attribute values.

\subsection{Benchmarking CBMs on \datasetname}
We report each model's performance in detecting the substituted target attribute ($\mathcal{S}^+$) in our synthetic \datasetname data, as presented in Table~\ref{tab:cbm_eval}.
$\mathcal{S}^+$ evaluates the 17 concepts in the overlap between attributes in \datasetname (33 attributes) and 112 (of 312) attributes used for CBM training \cite{cbm}.
We also measure the models' ability to remove the original attribute ($\mathcal{S}^-$);  100\% accuracy means that the CBM never predicted the removed attribute in the \datasetname data. For comparison with the accuracy of the classes seen in training, we include the CUB~\cite{cub} test set on the subset of attributes used in \datasetname ($\mathcal{T}_A$), and the test set accuracy on all CUB concepts ($\mathcal{T}$).

We observe that annotators consistently classify \datasetname attributes, achieving 94.0\% accuracy on the target attribute $\mathcal{S}^+$, and 96.8\% accuracy on not picking the removed attribute $\mathcal{S}^-$. These accuracies are much higher than CUB's (79.4\% for $\mathcal{T}_A$ and 82.5\% for $\mathcal{T}$). Despite all CBMs having high accuracy in $\mathcal{T}_A$ and $\mathcal{T}$ (up to 96.7\%), they generalize poorly to our novel attribute combinations (highest $\mathcal{S}^+$: $45.7\%$ from CEM~\cite{Zarlenga2022ConceptEM}). In fact, all tested CBMs detect $\mathcal{S}^+$ less accurately than random chance ($50\%$). %
Also, high $\mathcal{S}^-$ removal accuracy is often paired with low $\mathcal{S}^+$
(e.g. per-class soft labels with $91.6\%$ on $\mathcal{S}^-$, but $11.2\%$ on $\mathcal{S}^+$),
suggesting that some models %
may simply have a greater tendency to predict \textit{false} overall.
Among all tested CBMs, the CEM is the best at predicting $\mathcal{S}^+$ with $45.7\%$, albeit still below chance, while maintaining relatively high performance of $87.2\%$ on $\mathcal{S}^-$. %
All tested CBMs generalize poorly, showing that their predictions are not grounded in the target concepts and the performance on the training classes is misleading about the models' true interpretability. 

\subsection{Benchmarking Other VLMs on \datasetname}
Many interpretable models rely on VLM backbones such as CLIP~\cite{Radford2021LearningTV} to generalize to open-vocabulary settings without explicit training data~\cite{Rao2024DiscoverthenNameTC,menon2022visual,Oikarinen2023LabelFreeCB,yang2023language}. 
If these backbones also poorly ground predictions, all downstream models will be affected. In testing VLMs, we evaluate whether large-scale training can mitigate the challenges CBMs face and recognize \datasetname attributes, exhibited in Table~\ref{tab:cbm_eval:clip}.

As our tested VLMs cannot easily make binary predictions for individual attributes, we classify the target attribute among all within the attribute group or \textit{none}. The attribute with the highest cosine similarity compared to the image is labeled \textit{true} while all others are predicted \textit{false}. %
These open-set vocabulary models choose from the 312 CUB attributes~\cite{cub}, and we evaluate on the cleaner CBM class-aggregated labels~\cite{cbm}. We re-calculate random and human baselines given this multiclass classification setting (see Appendix F).%

Even with large-scale pre-training, CLIP~\cite{Radford2021LearningTV}, SigLIP~\cite{zhai2023sigmoid}, and EVA-CLIP~\cite{sun2023eva} continue to face challenges consistently identifying $\mathcal{S}^+$. While their overall accuracy is higher (with EVA-CLIP~\cite{sun2023eva} achieving $46.8\%$), a deeper look reveals interesting patterns. For instance, CLIP, SigLIP, and EVA-CLIP demonstrate a tendency to select the original attribute two to three times more often than random chance ($9.3\%$). This suggests a form of hallucination, where the models incorrectly identify the original attribute even when it is not present. For example, SigLIP 400m/16~\cite{zhai2023sigmoid}, the model with the lowest $\mathcal{S}^-$ hallucination rate, achieves only $81.7\%$, incorrectly selecting the original attribute $18.3\%$ of the time. %
This analysis suggests that despite advances in large-scale pre-training, the problem of generalization for individual concept classification remains a persistent hurdle for VLMs.

\begin{table}[]
    \centering
    \begin{tabular}{c c c|c c|c c}
         & \multirow{2}{*}{\rotatebox{90}{Per-Img.}} & \multirow{2}{*}[-1.65em]{\rotatebox{90}{Soft}} & \multicolumn{2}{c}{\datasetname} & \multicolumn{2}{c}{CUB} \\ [14pt]
         &  &  & $\mathcal{S}^+$  & $\mathcal{S}^-$ & $\mathcal{T}_A$ & $\mathcal{T}$ \\
         \hline \hline
         \textit{Random Chance} & & & \textit{50.0} & \textit{50.0} & \textit{50.0} & \textit{50.0} \\
         CBM (ind.)~\cite{cbm} & & & 40.8 & 51.6 & 95.9 & 96.7 \\
         CBM (joint)~\cite{cbm} & & & 34.3 & 54.0 & 96.1 & 96.9 \\
         CBM (per c.)~\cite{Heidemann2023ConceptCA} & & & 0.39 & 100.0 & 78.3 & 79.4 \\
         CEM~\cite{Zarlenga2022ConceptEM} & & & 45.7 & 87.2 & 77.9 & 81.3  \\
         \hline          
         CBM (ind.)~\cite{cbm} &  & \checkmark & 12.9 & 94.0 & 84.8 & 85.5 \\
         CBM (joint)~\cite{cbm} &  & \checkmark & 5.78 & 94.0 & 85.8 & 86.1 \\
         \hline
         CBM (ind.)~\cite{cbm} & \checkmark &  & 32.8 & 73.6 & 81.6 & 85.3 \\
         CBM (joint)~\cite{cbm} & \checkmark & & 27.1 & 67.1 & 83.4 & 86.0 \\
         \hline
         CBM (ind.)~\cite{cbm} & \checkmark & \checkmark & 11.2 & 91.6 & 80.9 & 82.6 \\
         CBM (joint)~\cite{cbm} & \checkmark & \checkmark & 6.06 & 92.4 & 74.9 & 75.8 \\
         \hline \hline
         Human & & & 94.0 & 96.8 & 79.4 & 82.5  \\
        
    \end{tabular}
    \caption{We evaluate CBM accuracy on \datasetname by measuring the substituted attribute ($\mathcal{S}^+$) and the removed attribute ($\mathcal{S}^-$), as well as on CUB ($\mathcal{T}$) and the SUB attribute label subset ($\mathcal{T}_A$).}%
    \label{tab:cbm_eval}
\end{table}

\begin{table}[t]
\centering
\begin{tabular}{c|c c|c c}
&\multicolumn{2}{c}{\datasetname} & \multicolumn{2}{c}{CUB} \\
& $\mathcal{S}^+$ & $\mathcal{S}^-$ & $\mathcal{T}_A$ & $\mathcal{T}$ \\
\hline \hline
\textit{Random Chance} & 9.3 & 90.7 & 73.3 & 74.6 \\
CLIP ViT-B32~\cite{Radford2021LearningTV} & 39.2 & 73.1 & 78.4 & 79.7 \\
CLIP ViT-L14~\cite{Radford2021LearningTV} & 45.5 & 73.2 & 78.6 & 80.1 \\
SigLIP B/16~\cite{zhai2023sigmoid} & 45.2 & 77.5 & 78.4 & 79.2 \\
SigLIP 400m/16~\cite{zhai2023sigmoid} & 45.7 & 81.7 & 77.6 & 79.2 \\
SigLIP2 B/16~\cite{tschannen2025siglip} & 40.0 & 76.6 & 77.6 & 79.1 \\
EVA-CLIP~\cite{sun2023eva} & 46.8 & 77.6 & 78.5 & 79.4 \\
\hline \hline
Human & 92.4 & 97.3 & 69.3 & 65.3  \\
\end{tabular}
\caption{We evaluate %
VLM accuracy on SUB by measuring the substituted attribute ($\mathcal{S}^+$) and removed attribute ($\mathcal{S}^-$), along with CUB accuracy ($\mathcal{T}$) and \datasetname's attribute subset ($\mathcal{T}_A$).}%
\label{tab:cbm_eval:clip}
\end{table}

\section{Limitations}
Since \method requires some human intervention for prompt creation and filtering verification, future improvements could focus on fully automating these processes. This would enable the creation of more explainable datasets, addressing the limited scope of CUB. %
Additionally, due to the automated filtering system, we cannot guarantee that every image in \datasetname perfectly represents the target attribute and bird. However, our human validation strongly suggests that \datasetname is more consistently and accurately labeled than CUB. Lastly, although our human studies involve fewer participants and annotations compared to CUB, we believe that the careful automatic filtering makes validating a subset sufficient as opposed to labeling the complete dataset.

\section{Conclusion}
\label{sec:conclusion}

We proposed a new dataset, \datasetname, to %
benchmark the grounding of attribute predictions in concept models. \datasetname consists of 38,400 images representing 768 unique bird-attribute pairs, where the given attribute is applied to the chosen bird. To generate \datasetname, we proposed \method, a test-time adaptation to faithfully generate novel attribute-object combinations by tying them to a second, easier image. Through rigorous filtering of the birds, attributes, and resulting images, we ensured the quality of \datasetname. We demonstrated that CBM concept predictions are poorly grounded in the target concepts and fail to generalize to unseen combinations of known concepts. We also revealed that SOTA VLMs experience this issue as well, though to a lesser extent. We hope that \datasetname will pave the way for the next generation of CBMs with more robust and well-grounded explanations.

\section*{Acknowledgments}
This work was partially funded by the ERC (853489 - DEXIM) and the Alfried Krupp von Bohlen und Halbach Foundation, which we thank for their generous support. The authors gratefully acknowledge the scientific support and resources of the AI service infrastructure \textit{LRZ AI Systems} provided by the Leibniz Supercomputing Centre (LRZ) of the Bavarian Academy of Sciences and Humanities (BAdW), funded by Bayerisches Staatsministerium für Wissenschaft und Kunst (StMWK).

{
    \small
    \bibliographystyle{ieeenat_fullname}
    \bibliography{main}

\begin{thebibliography}{72}
\providecommand{\natexlab}[1]{#1}
\providecommand{\url}[1]{\texttt{#1}}
\expandafter\ifx\csname urlstyle\endcsname\relax
  \providecommand{\doi}[1]{doi: #1}\else
  \providecommand{\doi}{doi: \begingroup \urlstyle{rm}\Url}\fi

\bibitem[Alam et~al.(2024)Alam, Srivastav, Kadir, and Sonntag]{Alam2024TowardsIR}
Hasan Md~Tusfiqur Alam, Devansh Srivastav, Md~Abdul Kadir, and Daniel Sonntag.
\newblock Towards interpretable radiology report generation via concept bottlenecks using a multi-agentic rag.
\newblock In \emph{arXiv}, 2024.

\bibitem[Alvarez-Melis and Jaakkola(2018)]{AlvarezMelis2018TowardsRI}
David Alvarez-Melis and T. Jaakkola.
\newblock Towards robust interpretability with self-explaining neural networks.
\newblock In \emph{NeurIPS}, 2018.

\bibitem[Aniraj et~al.(2024)Aniraj, Dantas, Ienco, and Marcos]{aniraj2024pdiscoformer}
Ananthu Aniraj, Cassio~F Dantas, Dino Ienco, and Diego Marcos.
\newblock Pdiscoformer: Relaxing part discovery constraints with vision transformers.
\newblock In \emph{ECCV}, 2024.

\bibitem[Brack et~al.(2024)Brack, Friedrich, Hintersdorf, Struppek, Schramowski, and Kersting]{sega}
Manuel Brack, Felix Friedrich, Dominik Hintersdorf, Lukas Struppek, Patrick Schramowski, and Kristian Kersting.
\newblock Sega: Instructing text-to-image models using semantic guidance.
\newblock In \emph{NeurIPS}, 2024.

\bibitem[Chen et~al.(2018)Chen, Li, Barnett, Su, and Rudin]{Chen2018ThisLL}
Chaofan Chen, Oscar Li, Alina~Jade Barnett, Jonathan Su, and Cynthia Rudin.
\newblock This looks like that: deep learning for interpretable image recognition.
\newblock In \emph{NeurIPS}, 2018.

\bibitem[Chen et~al.(2024{\natexlab{a}})Chen, Gao, Zhou, Wang, Li, Ge, and Zheng]{Chen2024EnhancingPF}
Hongyu Chen, Yi-Meng Gao, Min Zhou, Peng Wang, Xubin Li, Tiezheng Ge, and Bo Zheng.
\newblock Enhancing prompt following with visual control through training-free mask-guided diffusion.
\newblock In \emph{arXiv}, 2024{\natexlab{a}}.

\bibitem[Chen et~al.(2024{\natexlab{b}})Chen, Wang, Cao, Liu, Gao, Cui, Zhu, Ye, Tian, Liu, et~al.]{chen2024expanding}
Zhe Chen, Weiyun Wang, Yue Cao, Yangzhou Liu, Zhangwei Gao, Erfei Cui, Jinguo Zhu, Shenglong Ye, Hao Tian, Zhaoyang Liu, et~al.
\newblock Expanding performance boundaries of open-source multimodal models with model, data, and test-time scaling.
\newblock In \emph{arXiv}, 2024{\natexlab{b}}.

\bibitem[Chen et~al.(2024{\natexlab{c}})Chen, Wang, Tian, Ye, Gao, Cui, Tong, Hu, Luo, Ma, et~al.]{chen2024far}
Zhe Chen, Weiyun Wang, Hao Tian, Shenglong Ye, Zhangwei Gao, Erfei Cui, Wenwen Tong, Kongzhi Hu, Jiapeng Luo, Zheng Ma, et~al.
\newblock How far are we to gpt-4v? closing the gap to commercial multimodal models with open-source suites.
\newblock In \emph{arXiv}, 2024{\natexlab{c}}.

\bibitem[Chen et~al.(2024{\natexlab{d}})Chen, Wu, Wang, Su, Chen, Xing, Zhong, Zhang, Zhu, Lu, et~al.]{chen2024internvl}
Zhe Chen, Jiannan Wu, Wenhai Wang, Weijie Su, Guo Chen, Sen Xing, Muyan Zhong, Qinglong Zhang, Xizhou Zhu, Lewei Lu, et~al.
\newblock Internvl: Scaling up vision foundation models and aligning for generic visual-linguistic tasks.
\newblock In \emph{CVPR}, 2024{\natexlab{d}}.

\bibitem[Chowdhury et~al.(2024)Chowdhury, Phan, Liao, To, Xie, van~den Hengel, Verjans, and Liao]{Chowdhury2024AdaCBMAA}
Townim~Faisal Chowdhury, Vu~Minh~Hieu Phan, Kewen Liao, Minh-Son To, Yutong Xie, Anton van~den Hengel, Johan~W. Verjans, and Zhibin Liao.
\newblock Adacbm: An adaptive concept bottleneck model for explainable and accurate diagnosis.
\newblock In \emph{International Conference on Medical Image Computing and Computer-Assisted Intervention}, 2024.

\bibitem[Cunningham et~al.(2023)Cunningham, Ewart, Riggs, Huben, and Sharkey]{Cunningham2023SparseAF}
Hoagy Cunningham, Aidan Ewart, Logan Riggs, Robert Huben, and Lee Sharkey.
\newblock Sparse autoencoders find highly interpretable features in language models.
\newblock In \emph{arXiv}, 2023.

\bibitem[Dosovitskiy et~al.(2015)Dosovitskiy, Fischer, Ilg, H{\"a}usser, Hazirbas, Golkov, van~der Smagt, Cremers, and Brox]{Dosovitskiy2015FlowNetLO}
Alexey Dosovitskiy, Philipp Fischer, Eddy Ilg, Philip H{\"a}usser, Caner Hazirbas, Vladimir Golkov, Patrick van~der Smagt, Daniel Cremers, and Thomas Brox.
\newblock Flownet: Learning optical flow with convolutional networks.
\newblock In \emph{ICCV}, 2015.

\bibitem[Du and Kaelbling(2024)]{Du2024PositionCG}
Yilun Du and Leslie~Pack Kaelbling.
\newblock Position: Compositional generative modeling: A single model is not all you need.
\newblock In \emph{ICML}, 2024.

\bibitem[Du et~al.(2020)Du, Li, and Mordatch]{Du2020CompositionalVG}
Yilun Du, Shuang Li, and Igor Mordatch.
\newblock Compositional visual generation with energy based models.
\newblock In \emph{NeurIPS}, 2020.

\bibitem[Du et~al.(2023)Du, Durkan, Strudel, Tenenbaum, Dieleman, Fergus, Sohl-Dickstein, Doucet, and Grathwohl]{Du2023ReduceRR}
Yilun Du, Conor Durkan, Robin Strudel, Joshua~B. Tenenbaum, Sander Dieleman, Rob Fergus, Jascha~Narain Sohl-Dickstein, A. Doucet, and Will Grathwohl.
\newblock Reduce, reuse, recycle: Compositional generation with energy-based diffusion models and mcmc.
\newblock In \emph{ICML}, 2023.

\bibitem[Dunlap et~al.(2023)Dunlap, Umino, Zhang, Yang, Gonzalez, and Darrell]{Dunlap2023DiversifyYV}
Lisa Dunlap, Alyssa Umino, Han Zhang, Jiezhi Yang, Joseph~E. Gonzalez, and Trevor Darrell.
\newblock Diversify your vision datasets with automatic diffusion-based augmentation.
\newblock In \emph{NeurIPS}, 2023.

\bibitem[Eyring et~al.(2024)Eyring, Karthik, Roth, Dosovitskiy, and Akata]{Eyring2024ReNOEO}
Luca~Vincent Eyring, Shyamgopal Karthik, Karsten Roth, Alexey Dosovitskiy, and Zeynep Akata.
\newblock Reno: Enhancing one-step text-to-image models through reward-based noise optimization.
\newblock In \emph{NeurIPS}, 2024.

\bibitem[Feng et~al.(2024)Feng, Gong, Chen, Shen, Liu, and Zhou]{Feng2023RanniTT}
Yutong Feng, Biao Gong, Di Chen, Yujun Shen, Yu Liu, and Jingren Zhou.
\newblock Ranni: Taming text-to-image diffusion for accurate instruction following.
\newblock In \emph{CVPR}, 2024.

\bibitem[Furby et~al.(2024)Furby, Cunnington, Braines, and Preece]{Furby2024CanWC}
Jack Furby, Daniel Cunnington, Dave Braines, and Alun~David Preece.
\newblock Can we constrain concept bottleneck models to learn semantically meaningful input features?
\newblock In \emph{arXiv}, 2024.

\bibitem[Gaudi et~al.(2025)Gaudi, Sreekumar, and Boddeti]{Gaudi2025CoInDEL}
Sachit Gaudi, Gautam Sreekumar, and Vishnu~Naresh Boddeti.
\newblock Coind: Enabling logical compositions in diffusion models.
\newblock \emph{ICLR}, abs/2503.01145, 2025.

\bibitem[Havasi et~al.(2022)Havasi, Parbhoo, and Doshi-Velez]{Havasi2022AddressingLI}
Marton Havasi, S. Parbhoo, and Finale Doshi-Velez.
\newblock Addressing leakage in concept bottleneck models.
\newblock In \emph{NeurIPS}, 2022.

\bibitem[He et~al.(2023)He, Sun, Yu, Xue, Zhang, Torr, Bai, and Qi]{He2022IsSD}
Ruifei He, Shuyang Sun, Xin Yu, Chuhui Xue, Wenqing Zhang, Philip H.~S. Torr, Song Bai, and Xiaojuan Qi.
\newblock Is synthetic data from generative models ready for image recognition?
\newblock In \emph{ICLR}, 2023.

\bibitem[Heidemann et~al.(2023)Heidemann, Monnet, and Roscher]{Heidemann2023ConceptCA}
Lena Heidemann, Maureen Monnet, and Karsten Roscher.
\newblock Concept correlation and its effects on concept-based models.
\newblock In \emph{WACV}, 2023.

\bibitem[Hesse et~al.(2023)Hesse, Schaub-Meyer, and Roth]{Hesse2023FunnyBirdsAS}
Robin Hesse, Simone Schaub-Meyer, and Stefan Roth.
\newblock Funnybirds: A synthetic vision dataset for a part-based analysis of explainable ai methods.
\newblock In \emph{ICCV}, 2023.

\bibitem[Ho and Salimans(2021)]{ho2022classifier}
Jonathan Ho and Tim Salimans.
\newblock Classifier-free diffusion guidance.
\newblock In \emph{NeurIPS Workshop on Deep Generative Models and Downstream Applications}, 2021.

\bibitem[Horn et~al.(2015)Horn, Branson, Farrell, Haber, Barry, Ipeirotis, Perona, and Belongie]{Horn2015BuildingAB}
Grant~Van Horn, Steve Branson, Ryan Farrell, Scott Haber, Jessie Barry, Panagiotis~G. Ipeirotis, Pietro Perona, and Serge~J. Belongie.
\newblock Building a bird recognition app and large scale dataset with citizen scientists: The fine print in fine-grained dataset collection.
\newblock In \emph{CVPR}, 2015.

\bibitem[Johnson et~al.(2017)Johnson, Hariharan, van~der Maaten, Fei-Fei, Zitnick, and Girshick]{Johnson2016CLEVRAD}
Justin Johnson, Bharath Hariharan, Laurens van~der Maaten, Li Fei-Fei, C.~Lawrence Zitnick, and Ross~B. Girshick.
\newblock Clevr: A diagnostic dataset for compositional language and elementary visual reasoning.
\newblock In \emph{CVPR}, 2017.

\bibitem[Kim et~al.(2024{\natexlab{a}})Kim, Thomas, and Ghadiyaram]{Kim2024RevelioIA}
Dahye Kim, Xavier Thomas, and Deepti Ghadiyaram.
\newblock Revelio: Interpreting and leveraging semantic information in diffusion models.
\newblock In \emph{arXiv}, 2024{\natexlab{a}}.

\bibitem[Kim et~al.(2024{\natexlab{b}})Kim, Bader, Alaniz, Schmid, and Akata]{Kim2024DataDreamFG}
Jae~Myung Kim, Jessica Bader, Stephan Alaniz, Cordelia Schmid, and Zeynep Akata.
\newblock Datadream: Few-shot guided dataset generation.
\newblock In \emph{ECCV}, 2024{\natexlab{b}}.

\bibitem[Koh et~al.(2020)Koh, Nguyen, Tang, Mussmann, Pierson, Kim, and Liang]{cbm}
Pang~Wei Koh, Thao Nguyen, Yew~Siang Tang, Stephen Mussmann, Emma Pierson, Been Kim, and Percy Liang.
\newblock Concept bottleneck models.
\newblock In \emph{ICML}, 2020.

\bibitem[Labs(2024)]{flux}
Black~Forest Labs.
\newblock Announcing black forest labs.
\newblock In \emph{BlackForestLabs Blog}, 2024.

\bibitem[Liu et~al.(2021)Liu, Li, Du, Tenenbaum, and Torralba]{Liu2021LearningTC}
Nan Liu, Shuang Li, Yilun Du, Joshua~B. Tenenbaum, and Antonio Torralba.
\newblock Learning to compose visual relations.
\newblock \emph{NeurIPS}, 2021.

\bibitem[Liu et~al.(2022)Liu, Li, Du, Torralba, and Tenenbaum]{Liu2022CompositionalVG}
Nan Liu, Shuang Li, Yilun Du, Antonio Torralba, and Joshua~B. Tenenbaum.
\newblock Compositional visual generation with composable diffusion models.
\newblock \emph{ECCV}, 2022.

\bibitem[Mahinpei et~al.(2021)Mahinpei, Clark, Lage, Doshi-Velez, and Pan]{Mahinpei2021PromisesAP}
Anita Mahinpei, Justin Clark, Isaac Lage, Finale Doshi-Velez, and Weiwei Pan.
\newblock Promises and pitfalls of black-box concept learning models.
\newblock In \emph{ArXiv}, 2021.

\bibitem[Makhzani and Frey(2013)]{Makhzani2013kSparseA}
Alireza Makhzani and Brendan~J. Frey.
\newblock k-sparse autoencoders.
\newblock In \emph{arXiv}, 2013.

\bibitem[Marconato et~al.(2022)Marconato, Passerini, and Teso]{Marconato2022GlanceNetsIL}
Emanuele Marconato, Andrea Passerini, and Stefano Teso.
\newblock Glancenets: Interpretabile, leak-proof concept-based models.
\newblock In \emph{NeurIPS}, 2022.

\bibitem[Margeloiu et~al.(2021)Margeloiu, Ashman, Bhatt, Chen, Jamnik, and Weller]{Margeloiu2021DoCB}
Andrei Margeloiu, Matthew Ashman, Umang Bhatt, Yanzhi Chen, Mateja Jamnik, and Adrian Weller.
\newblock Do concept bottleneck models learn as intended?
\newblock In \emph{arXiv}, 2021.

\bibitem[Menon and Vondrick(2023)]{menon2022visual}
Sachit Menon and Carl Vondrick.
\newblock Visual classification via description from large language models.
\newblock \emph{ICLR}, 2023.

\bibitem[Mo et~al.(2024)Mo, Mu, Lin, Liu, Guan, Li, and Zhou]{Mo2023FreeControlTS}
Sicheng Mo, Fangzhou Mu, Kuan~Heng Lin, Yanli Liu, Bochen Guan, Yin Li, and Bolei Zhou.
\newblock Freecontrol: Training-free spatial control of any text-to-image diffusion model with any condition.
\newblock In \emph{CVPR}, 2024.

\bibitem[Nauta et~al.(2023)Nauta, Schl{\"o}tterer, van Keulen, and Seifert]{Nauta2023PIPNetPI}
Meike Nauta, J{\"o}rg Schl{\"o}tterer, Maurice van Keulen, and Christin Seifert.
\newblock Pip-net: Patch-based intuitive prototypes for interpretable image classification.
\newblock In \emph{CVPR}, 2023.

\bibitem[Nnamdi et~al.(2023)Nnamdi, Shi, Tamo, Iwinski, Wattenbarger, and Wang]{Nnamdi2023ConceptBM}
Micky~C. Nnamdi, Wenqi Shi, Junior~Ben Tamo, Henry~J. Iwinski, J.~Michael Wattenbarger, and May~Dongmei Wang.
\newblock Concept bottleneck model for adolescent idiopathic scoliosis patient reported outcomes prediction.
\newblock In \emph{International Conference on Biomedical and Health Informatics (BHI)}, 2023.

\bibitem[Oikarinen et~al.(2023)Oikarinen, Das, Nguyen, and Weng]{Oikarinen2023LabelFreeCB}
Tuomas~P. Oikarinen, Subhro Das, Lam~M. Nguyen, and Tsui-Wei Weng.
\newblock Label-free concept bottleneck models.
\newblock In \emph{ICLR}, 2023.

\bibitem[Pach et~al.(2024)Pach, Rymarczyk, Lewandowska, Tabor, and Zieliński]{Pach2024LucidPPNUP}
Mateusz Pach, Dawid Rymarczyk, Koryna Lewandowska, Jacek Tabor, and Bartosz Zieliński.
\newblock Lucidppn: Unambiguous prototypical parts network for user-centric interpretable computer vision.
\newblock In \emph{arXiv}, 2024.

\bibitem[Panousis et~al.(2024)Panousis, Ienco, and Marcos]{panousis2024coarse}
Konstantinos Panousis, Dino Ienco, and Diego Marcos.
\newblock Coarse-to-fine concept bottleneck models.
\newblock In \emph{NeurIPS}, 2024.

\bibitem[Park et~al.(2019)Park, Darrell, and Rohrbach]{Park2019RobustCC}
Dong~Huk Park, Trevor Darrell, and Anna Rohrbach.
\newblock Robust change captioning.
\newblock In \emph{ICCV}, 2019.

\bibitem[Podell et~al.(2024)Podell, English, Lacey, Blattmann, Dockhorn, Muller, Penna, and Rombach]{Podell2023SDXLIL}
Dustin Podell, Zion English, Kyle Lacey, A. Blattmann, Tim Dockhorn, Jonas Muller, Joe Penna, and Robin Rombach.
\newblock Sdxl: Improving latent diffusion models for high-resolution image synthesis.
\newblock In \emph{ICLR}, 2024.

\bibitem[Radford et~al.(2021)Radford, Kim, Hallacy, Ramesh, Goh, Agarwal, Sastry, Askell, Mishkin, Clark, Krueger, and Sutskever]{Radford2021LearningTV}
Alec Radford, Jong~Wook Kim, Chris Hallacy, Aditya Ramesh, Gabriel Goh, Sandhini Agarwal, Girish Sastry, Amanda Askell, Pamela Mishkin, Jack Clark, Gretchen Krueger, and Ilya Sutskever.
\newblock Learning transferable visual models from natural language supervision.
\newblock In \emph{ICML}, 2021.

\bibitem[Raman et~al.(2024{\natexlab{a}})Raman, Zarlenga, Heo, and Jamnik]{Raman2024DoCB}
Naveen Raman, Mateo~Espinosa Zarlenga, Juyeon Heo, and Mateja Jamnik.
\newblock Do concept bottleneck models respect localities?
\newblock In \emph{arXiv}, 2024{\natexlab{a}}.

\bibitem[Raman et~al.(2024{\natexlab{b}})Raman, Zarlenga, and Jamnik]{Raman2024UnderstandingIR}
Naveen Raman, Mateo~Espinosa Zarlenga, and Mateja Jamnik.
\newblock Understanding inter-concept relationships in concept-based models.
\newblock In \emph{ICML}, 2024{\natexlab{b}}.

\bibitem[Rao et~al.(2024)Rao, Mahajan, Bohle, and Schiele]{Rao2024DiscoverthenNameTC}
Sukrut Rao, Sweta Mahajan, Moritz Bohle, and Bernt Schiele.
\newblock Discover-then-name: Task-agnostic concept bottlenecks via automated concept discovery.
\newblock In \emph{ECCV}, 2024.

\bibitem[Rombach et~al.(2022)Rombach, Blattmann, Lorenz, Esser, and Ommer]{Rombach2021HighResolutionIS}
Robin Rombach, A. Blattmann, Dominik Lorenz, Patrick Esser, and Bj{\"o}rn Ommer.
\newblock High-resolution image synthesis with latent diffusion models.
\newblock In \emph{CVPR}, 2022.

\bibitem[Rymarczyk et~al.(2020)Rymarczyk, Struski, Tabor, and Zieliński]{Rymarczyk2020ProtoPSharePP}
Dawid Rymarczyk, Lukasz Struski, Jacek Tabor, and Bartosz Zieliński.
\newblock Protopshare: Prototypical parts sharing for similarity discovery in interpretable image classification.
\newblock In \emph{ACM SIGKDD Conference on Knowledge Discovery \& Data Mining}, 2020.

\bibitem[Rymarczyk et~al.(2021)Rymarczyk, Struski, G'orszczak, Lewandowska, Tabor, and Zieliński]{Rymarczyk2021InterpretableIC}
Dawid Rymarczyk, Lukasz Struski, Michal G'orszczak, Koryna Lewandowska, Jacek Tabor, and Bartosz Zieliński.
\newblock Interpretable image classification with differentiable prototypes assignment.
\newblock In \emph{ECCV}, 2021.

\bibitem[Sariyildiz et~al.(2022)Sariyildiz, Karteek, Larlus, and Kalantidis]{Sariyildiz2022FakeIT}
Mert~Bulent Sariyildiz, Alahari Karteek, Diane Larlus, and Yannis Kalantidis.
\newblock Fake it till you make it: Learning transferable representations from synthetic imagenet clones.
\newblock In \emph{CVPR}, 2022.

\bibitem[Sawada and Nakamura(2022)]{Sawada2022ConceptBM}
Yoshihide Sawada and Keigo Nakamura.
\newblock Concept bottleneck model with additional unsupervised concepts.
\newblock In \emph{IEEE Access}, 2022.

\bibitem[Singhi et~al.(2024)Singhi, Kim, Roth, and Akata]{Singhi2024ImprovingIE}
Nishad Singhi, Jae~Myung Kim, Karsten Roth, and Zeynep Akata.
\newblock Improving intervention efficacy via concept realignment in concept bottleneck models.
\newblock In \emph{ECCV}, 2024.

\bibitem[Sinha et~al.(2022)Sinha, Huai, Sun, and Zhang]{Sinha2022UnderstandingAE}
Sanchit Sinha, Mengdi Huai, Jianhui Sun, and Aidong Zhang.
\newblock Understanding and enhancing robustness of concept-based models.
\newblock In \emph{AAAI}, 2022.

\bibitem[Sun et~al.(2023)Sun, Fang, Wu, Wang, and Cao]{sun2023eva}
Quan Sun, Yuxin Fang, Ledell Wu, Xinlong Wang, and Yue Cao.
\newblock Eva-clip: Improved training techniques for clip at scale.
\newblock \emph{arXiv}, 2023.

\bibitem[Tan et~al.(2024)Tan, Zhou, and Chen]{Tan2024ExplainVA}
Andong Tan, Fengtao Zhou, and Hao Chen.
\newblock Explain via any concept: Concept bottleneck model with open vocabulary concepts.
\newblock In \emph{ECCV}, 2024.

\bibitem[Thasarathan et~al.(2025)Thasarathan, Forsyth, Fel, Kowal, and Derpanis]{Thasarathan2025UniversalSA}
Harrish Thasarathan, Julian Forsyth, Thomas Fel, Matthew Kowal, and Konstantinos Derpanis.
\newblock Universal sparse autoencoders: Interpretable cross-model concept alignment.
\newblock In \emph{arXiv}, 2025.

\bibitem[Tschannen et~al.(2025)Tschannen, Gritsenko, Wang, Naeem, Alabdulmohsin, Parthasarathy, Evans, Beyer, Xia, Mustafa, et~al.]{tschannen2025siglip}
Michael Tschannen, Alexey Gritsenko, Xiao Wang, Muhammad~Ferjad Naeem, Ibrahim Alabdulmohsin, Nikhil Parthasarathy, Talfan Evans, Lucas Beyer, Ye Xia, Basil Mustafa, et~al.
\newblock Siglip 2: Multilingual vision-language encoders with improved semantic understanding, localization, and dense features.
\newblock \emph{arXiv}, 2025.

\bibitem[van~der Klis et~al.(2023)van~der Klis, Alaniz, Mancini, Dantas, Ienco, Akata, and Marcos]{van2023pdisconet}
Robert van~der Klis, Stephan Alaniz, Massimiliano Mancini, Cassio~F Dantas, Dino Ienco, Zeynep Akata, and Diego Marcos.
\newblock Pdisconet: Semantically consistent part discovery for fine-grained recognition.
\newblock In \emph{ICCV}, 2023.

\bibitem[Wah et~al.(2011)Wah, Branson, Welinder, Perona, and Belongie]{cub}
C. Wah, S. Branson, P. Welinder, P. Perona, and S. Belongie.
\newblock The caltech-ucsd birds-200-2011 dataset.
\newblock In \emph{California Institute of Technology Technical Report}, 2011.

\bibitem[Wang et~al.(2024)Wang, Liu, Hsieh, and Gong]{Wang2024OnDP}
Ruochen Wang, Ting Liu, Cho-Jui Hsieh, and Boqing Gong.
\newblock On discrete prompt optimization for diffusion models.
\newblock In \emph{ICML}, 2024.

\bibitem[Wiedemer et~al.(2023)Wiedemer, Mayilvahanan, Bethge, and Brendel]{Wiedemer2023CompositionalGF}
Thadd{\"a}us Wiedemer, Prasanna Mayilvahanan, Matthias Bethge, and Wieland Brendel.
\newblock Compositional generalization from first principles.
\newblock \emph{NeurIPS}, 2023.

\bibitem[Yang et~al.(2023)Yang, Panagopoulou, Zhou, Jin, Callison-Burch, and Yatskar]{yang2023language}
Yue Yang, Artemis Panagopoulou, Shenghao Zhou, Daniel Jin, Chris Callison-Burch, and Mark Yatskar.
\newblock Language in a bottle: Language model guided concept bottlenecks for interpretable image classification.
\newblock In \emph{CVPR}, 2023.

\bibitem[Zarlenga et~al.(2022)Zarlenga, Barbiero, Ciravegna, Marra, Giannini, Diligenti, Shams, Precioso, Melacci, Weller, Lio', and Jamnik]{Zarlenga2022ConceptEM}
Mateo~Espinosa Zarlenga, Pietro Barbiero, Gabriele Ciravegna, Giuseppe Marra, Francesco Giannini, Michelangelo Diligenti, Zohreh Shams, Fr{\'e}d{\'e}ric Precioso, Stefano Melacci, Adrian Weller, Pietro Lio', and Mateja Jamnik.
\newblock Concept embedding models.
\newblock In \emph{NeurIPS}, 2022.

\bibitem[Zarlenga et~al.(2023)Zarlenga, Barbiero, Shams, Kazhdan, Bhatt, Weller, and Jamnik]{Zarlenga2023TowardsRM}
Mateo~Espinosa Zarlenga, Pietro Barbiero, Zohreh Shams, Dmitry Kazhdan, Umang Bhatt, Adrian Weller, and Mateja Jamnik.
\newblock Towards robust metrics for concept representation evaluation.
\newblock In \emph{AAAI}, 2023.

\bibitem[Zhai et~al.(2023)Zhai, Mustafa, Kolesnikov, and Beyer]{zhai2023sigmoid}
Xiaohua Zhai, Basil Mustafa, Alexander Kolesnikov, and Lucas Beyer.
\newblock Sigmoid loss for language image pre-training.
\newblock In \emph{ICCV}, 2023.

\bibitem[Zhang et~al.(2024)Zhang, Shen, Li, and Liu]{Zhang2024LargeMM}
Kaichen Zhang, Yifei Shen, Bo Li, and Ziwei Liu.
\newblock Large multi-modal models can interpret features in large multi-modal models.
\newblock In \emph{arXiv}, 2024.

\bibitem[Zhang et~al.(2023)Zhang, Rao, and Agrawala]{Zhang2023AddingCC}
Lvmin Zhang, Anyi Rao, and Maneesh Agrawala.
\newblock Adding conditional control to text-to-image diffusion models.
\newblock In \emph{ICCV}, 2023.

\bibitem[Zhao et~al.(2023)Zhao, Chen, Chen, Bao, Hao, Yuan, and Wong]{Zhao2023UniControlNetAC}
Shihao Zhao, Dongdong Chen, Yen-Chun Chen, Jianmin Bao, Shaozhe Hao, Lu Yuan, and Kwan-Yee~K. Wong.
\newblock Uni-controlnet: All-in-one control to text-to-image diffusion models.
\newblock In \emph{NeurIPS}, 2023.

\end{thebibliography}
}

\newpage \clearpage
\appendix
\section{Reference Birds}
\label{sec:reference_birds}
For \datasetname, we use the following 33 reference birds: Western Grebe, Black and white Warbler, European Goldfinch, Pacific Loon, White Pelican, Cedar Waxwing, Gadwall, Downy Woodpecker, Pileated Woodpecker, Purple Finch, Common Raven, White breasted Nuthatch, Northern Flicker, Mallard, Tropical Kingbird, Tree Swallow, Song Sparrow, Green Violetear, Gray Catbird, Green Jay, Cardinal, Red bellied Woodpecker, Pied Kingfisher, Rufous Hummingbird, Dark eyed Junco, Green Kingfisher, Horned Puffin, Anna Hummingbird, Barn Swallow, American Goldfinch, Lazuli Bunting, Blue Jay, Painted Bunting.

\section{Guidance Birds}
\label{sec:guidance_birds}
Guidance birds are used for pattern and shape modifications. We include in Table~\ref{tab:guidance_birds} the guidance birds chosen for each attribute when generating \datasetname.

\begin{table}[]
\begin{tabular}{ll}
Attribute                  & Guidance Bird      \\
\hline
Needle bill shape          & Needle bill shape  \\
Spotted breast pattern     & Brown Thrasher \\
Striped breast pattern     & Song Sparrow \\
Solid tail pattern         & Gray Catbird \\
Multi-colored tail pattern & Cedar Waxwing 
\end{tabular}
\caption{Guidance birds used the the generation of \datasetname.}
\label{tab:guidance_birds}
\end{table}

\section{Substitutions}
\label{sec:substitutions}
We use the following list of substitutions in \datasetname: grey back color, grey bill color, white breast color, red breast color, blue breast color, grey crown color, white crown color, black crown color, pink crown color, yellow eye color, blue eye color, white eye color, grey forehead color, pink leg color, black leg color, grey leg color, green primary color, brown primary color, blue primary color, orange primary color, blue throat color, yellow throat color, green underparts color, red underparts color, white wing color, grey wing color, black wing color, spotted breast pattern, striped breast pattern, solid tail pattern, multi-colored tail pattern, and needle bill shape. 

\section{Prompts}
\label{sec:prompt_examples}
A few example prompts:

$\mathcal{R} =$ European Goldfinch, $\mathcal{S}^+ =$ Black crown color, $\mathcal{G}=$ bird, $c_\mathcal{R} =$ A photo of a European Goldfinch with black colored feathers on the crown of its head, $c_\mathcal{G} =$ A photo of a bird with black colored feathers on the crown of its head 

$\mathcal{R} =$ Downy Woodpecker, $\mathcal{S}^+ =$ Red breast color, $\mathcal{G}=$ bird, $c_\mathcal{R} =$ A photo of a Downy Woodpecker with a red colored breast, $c_\mathcal{G} =$ A photo of a bird with a red colored breast 

$\mathcal{R} =$ Western Grebe, $\mathcal{S}^+ =$ Solid tail pattern, $\mathcal{G}=$ Gray Catbird, $c_\mathcal{R} =$ A photo of a Western Grebe with a solid tail like a Gray Catbird, $c_\mathcal{G} =$ A photo of a Gray Catbird with a solid tail 

$\mathcal{R} =$ Cardinal, $\mathcal{S}^+ =$ Spotted breast pattern, $\mathcal{G}=$ Brown Thrasher, $c_\mathcal{R} =$ A photo of a Cardinal with a  spotted belly like a Brown Thrasher, $c_\mathcal{G} =$ A photo of a Brown Thrasher with a spotted belly

$\mathcal{R} =$ Blue Jay, $\mathcal{S}^+ =$ Needle bill shape, $\mathcal{G}=$ Hummingbird, $c_\mathcal{R} =$ A photo of a Blue Jay with the body of a Blue Jay and a beak like a Hummingbird, $c_\mathcal{G} =$ A photo of a Hummingbird                                                        
\section{Human Verification User Interface}
\label{sec:human_verification}
\begin{figure}[t]
    \centering
    \includegraphics[width=\linewidth]{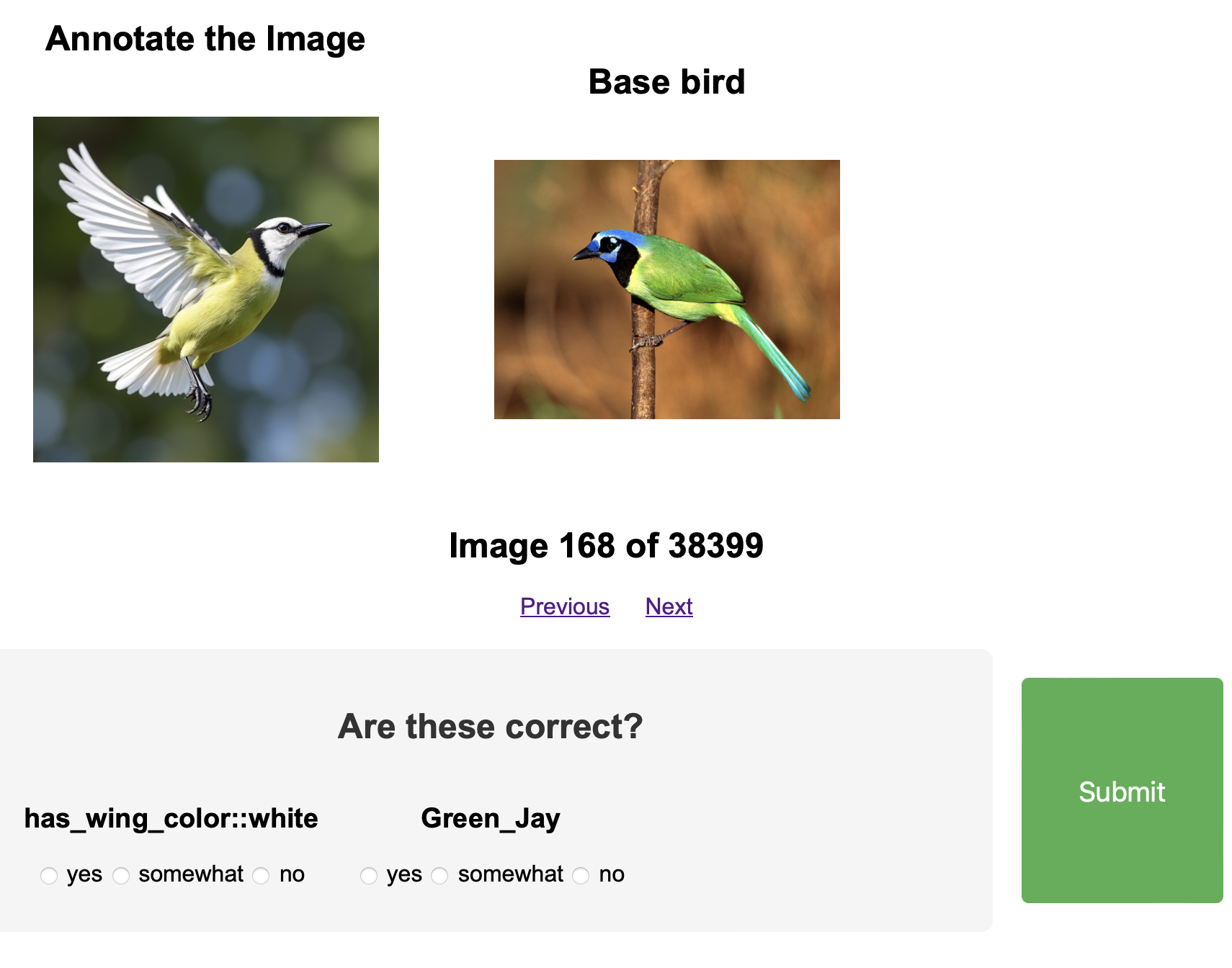} 
    \vspace{-10pt}
    \caption{
        First human study interface with binary questions for attribute presence and reference bird faithfulness.
    }
    \label{fig:human_study_interface}
\end{figure}

\begin{figure}[t]
    \centering
    \includegraphics[width=\linewidth]{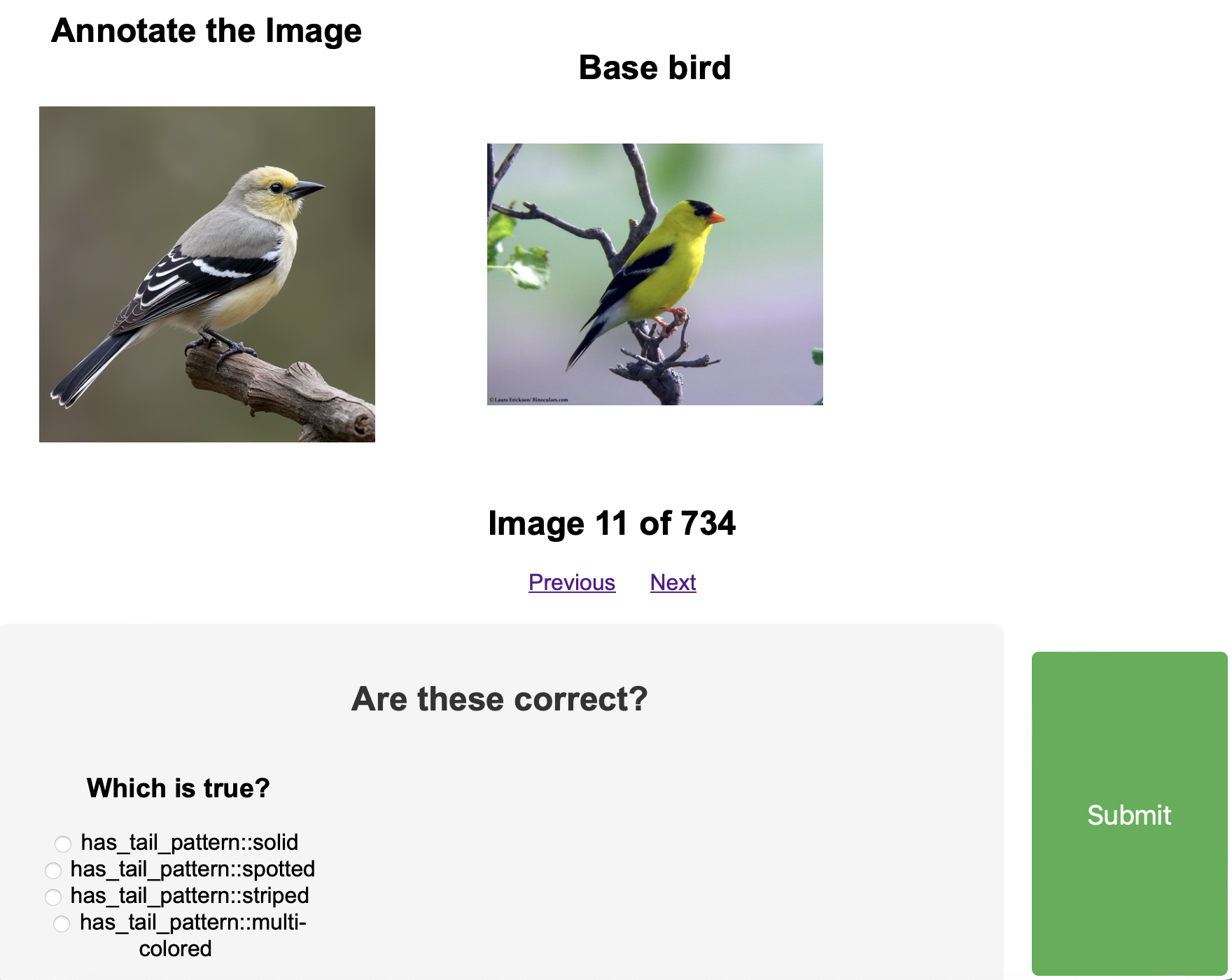} 
    \vspace{-10pt}
    \caption{
        Second human study with attribute labeling within full attribute group.
    }
    \label{fig:human_study_interface2}
\end{figure}

Human verification was completed by four volunteers. In Figure~\ref{fig:human_study_interface}, we see the user interface used for our first user study, where participants were asked whether $\mathcal{S}^+$ was present and whether the bird accurately reflected the guide bird. In Figure~\ref{fig:human_study_interface2}, we show the interface for the second study, where the user is given all options in the target attribute group and asked to label which is present.

\begin{figure}[t]
    \centering
    \includegraphics[width=\linewidth]{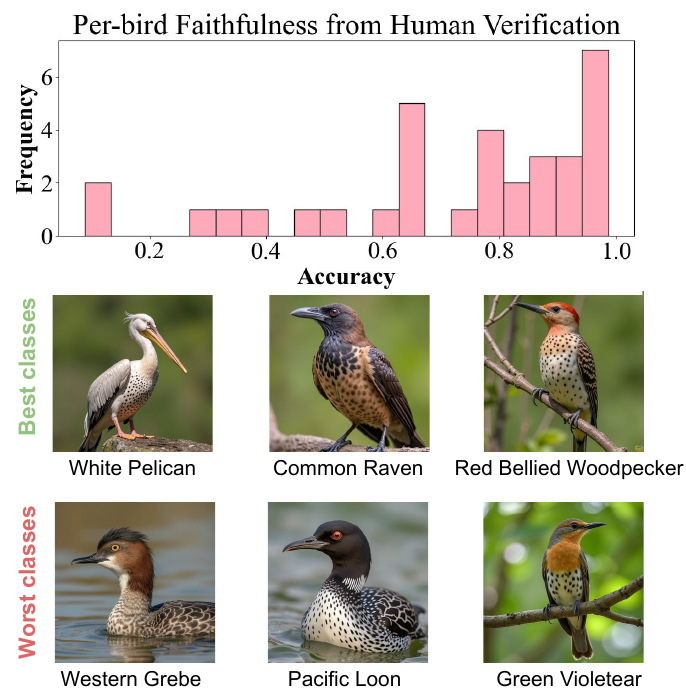} 
    \vspace{-10pt}
    \caption{
        Histogram of the per-bird accuracy results from the human verification, where participants were asked if the attribute-substituted synthetic image represents the original bird. We see that 27 birds exceed 50\% accuracy, showing that our generated dataset is very diverse.
    }
    \label{fig:bird_acc_hist}
\end{figure}

\subsection{Reference Bird Verification}
The underlying objective of specifying reference birds is to increase the overall diversity in birds exhibiting individual attributes. Specifically, we want to test the accuracy of attribute detection when it occurs in combinations not seen during test time. As long as $\mathcal{S}^+$ is present, it is not imperative that every synthetic bird closely match the reference class, but many should. As described in Section~\ref{sec:human_val}, we verify this on 40 images per attribute, by checking if the synthetic bird is recognizable as the reference bird. We then calculate the percentage of faithful birds of all those generated for each bird class. In Figure~\ref{fig:bird_acc_hist}, we show a histogram of this per-bird faithfulness. From this histogram, we can see that 27 out of 33 classes are faithful over half the time, and 10 classes are over 90\% faithful. For the attribute \textit{spotted breast}, we show examples from the three most faithful classes, and the three least faithful. While \textit{Western Grebe}, \textit{Pacific Loon}, and \textit{Green Violetear} diverge from the representative class, we also note that they still provide some diversity to \datasetname.

\section{VLM Random Chance Calculation and Label Set}
\label{sec:random_chance}
For the VLMs, we calculate the probability of getting a single prediction correct at random if the target label is 1 as $\frac{1}{|\mathcal{A}| + 1}$, where $\mathcal{A}$ is the attribute group corresponding to the target prediction and options $a_j \in \mathcal{A}$ are the manifestations of the attribute group. One is added to $|\mathcal{A}|$ to account for the additional option \textit{none}. If the target label is 0, then it is $1 - \frac{1}{|\mathcal{A}| + 1}$.

For \datasetname, we calculate the $\mathcal{S}^+$ random chance baseline across the modified attribute for each image in \datasetname, assuming a target label of 1. We calculate $\mathcal{S}^-$ from only the samples where the class-wise CBM label included a positive label for another attribute within the attribute group, and we consider that attribute with a target label of 0.

For CUB, we calculate the random chance baseline across all samples and CBM attributes, with the CBM class-wise labels as targets.

For selecting the possible label set $\mathcal{A}$ presented to the VLM, we use the full set of 312 CUB attributes for two reasons: (1) it offers a broader and more challenging set of plausible options than the CBM subset; and (2) the original labels used in CUB collection are well-aligned with expected dataset attributes, increasing the likelihood that the model selects the correct attribute over \textit{none}.

\end{document}